\documentclass[final]{cvpr}

\usepackage{times}
\usepackage{epsfig}
\usepackage{graphicx}
\usepackage{amsmath}
\usepackage{amssymb}

\usepackage{xcolor}
\usepackage{subcaption}
\usepackage{bm}
\usepackage{enumitem}
\usepackage{balance}

\usepackage[pagebackref=true,breaklinks=true,colorlinks,bookmarks=false]{hyperref}

\def\cvprPaperID{****} 
\def\confYear{CVPR 2021}
\begin{document}

\title{RfD-Net: Point Scene Understanding by Semantic Instance Reconstruction}

\author{Yinyu Nie\textsuperscript{1,2} \qquad Ji Hou\textsuperscript{3} \qquad Xiaoguang Han\textsuperscript{1,}$^{\ast}$ \qquad Matthias Nie{\ss}ner\textsuperscript{3}\\
	\textsuperscript{1}SRIBD, CUHKSZ \qquad
	\textsuperscript{2}Bournemouth University \qquad
	\textsuperscript{3}Technical University of Munich 
}

\twocolumn[{%
	\renewcommand\twocolumn[1][]{#1}%
	\maketitle
	\begin{center}
		\includegraphics[width=0.3\textwidth]{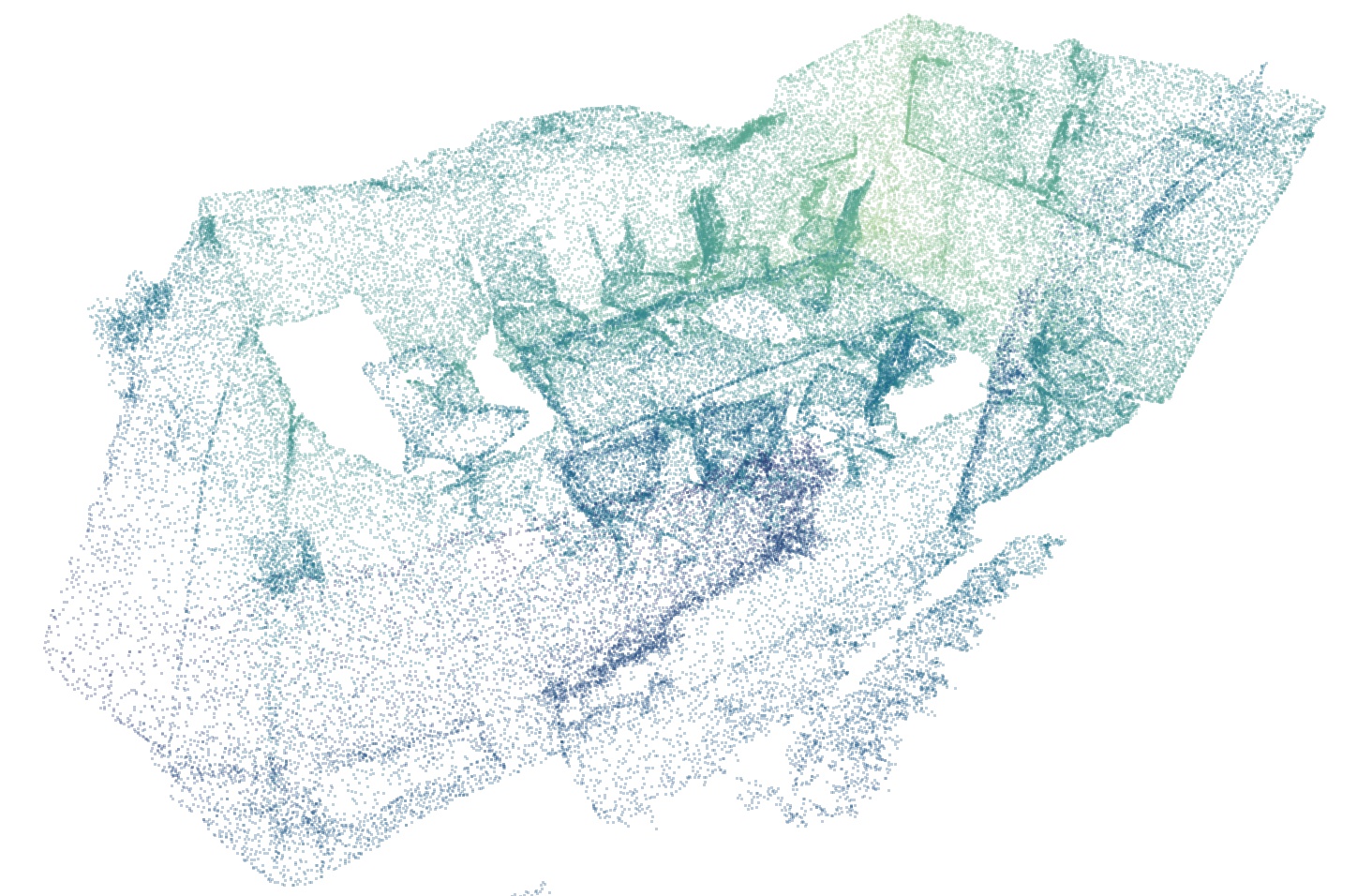}
		\includegraphics[width=0.3\textwidth]  
		{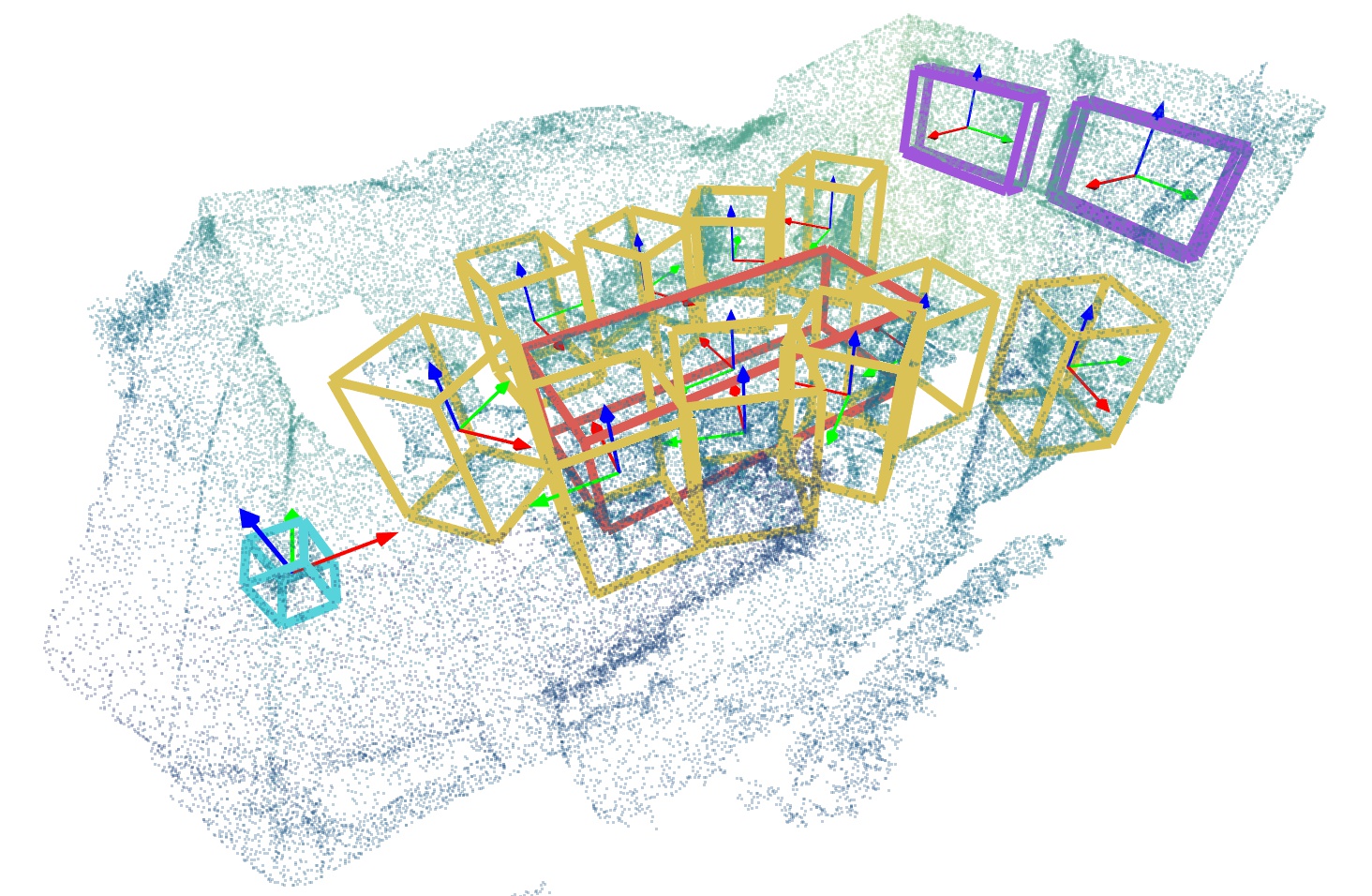}
		\includegraphics[width=0.3\textwidth]  
		{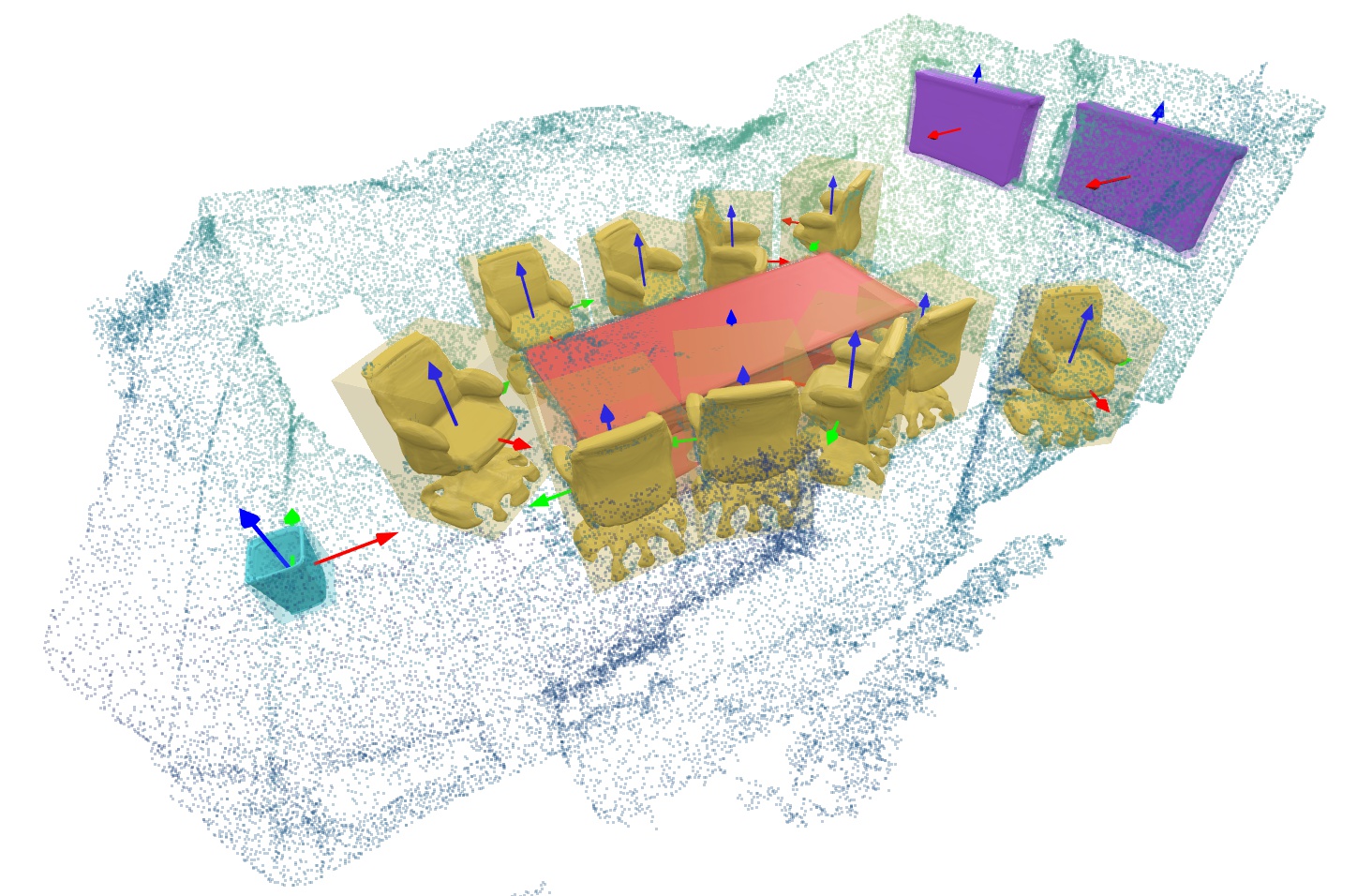}
		\captionof{figure}{From an incomplete point cloud ($N\times3$) of a 3D scene (left), our method learns to jointly understand the 3D objects with semantic labels, poses (middle) and complete object meshes (right).
		}
		\label{fig:teaser}
	\end{center}
}]

\let\thefootnote\relax\footnotetext{$^{\ast}$ Corresponding Email: {\tt\small hanxiaoguang@cuhk.edu.cn}}

\begin{abstract}
	Semantic scene understanding from point clouds is particularly challenging as the points reflect only a sparse set of the underlying 3D geometry. Previous works often convert point cloud into regular grids (e.g. voxels or bird-eye view images), and resort to grid-based convolutions for scene understanding. In this work, we introduce \textit{RfD-Net} that jointly detects and reconstructs dense object surfaces directly from raw point clouds. Instead of representing scenes with regular grids, our method leverages the sparsity of point cloud data and focuses on predicting shapes that are recognized with high objectness. With this design, we decouple the instance reconstruction into global object localization and local shape prediction. It not only eases the difficulty of learning 2-D manifold surfaces from sparse 3D space, the point clouds in each object proposal convey shape details that support implicit function learning to reconstruct any high-resolution surfaces. Our experiments indicate that instance detection and reconstruction present complementary effects, where the shape prediction head shows consistent effects on improving object detection with modern 3D proposal network backbones. The qualitative and quantitative evaluations further demonstrate that our approach consistently outperforms the state-of-the-arts and improves over 11 of mesh IoU in object reconstruction.
\end{abstract}
\vspace{-20pt}

\section{Introduction}
Semantic scene reconstruction has received increasing attention in applications such as robot navigation and interior design. It focuses on recovering the object labels, poses and geometries of objects in a 3D scene from partial observations (e.g. images or 3D scans). With the advance of 2D CNNs, instance reconstruction from images achieves appealing results \cite{gkioxari2019mesh,nie2020total3dunderstanding,li2020frodo,tulsiani2018factoring,kundu20183d} but still bottlenecked by the depth ambiguity thus resulting in defective object locations. Compared to images, point clouds provide surface geometry that largely alleviates the object locating issues \cite{qi2019deep,ahmed2020density,xie2020mlcvnet,chen2020hierarchical,jiang2020pointgroup}. However, its inherent sparseness and irregularity challenge the direct usage of grid-based CNNs on point clouds for semantic instance reconstruction.

Scanning 3D scenes usually results in missing geometries due to occlusions, view constraints and weak illumination where individual objects cannot be covered in all views. Prior works have explored various strategies to recover the missing shapes, e.g. depth inpainting, voxel/TSDF prediction, and shape retrieval. Depth inpainting \cite{han2019deep,novotny2019perspectivenet,huang2019indoor,senushkin2020decoder} aims to complete the depth maps in single views. With 2D CNNs, these methods achieve pleasing results in surface points recovery. To complete a scene with occluded contents, many methods extend the 2D CNNs to 3D and naturally represent 3D scenes with voxel/TSDF grids \cite{song2017semantic,wang2019forknet,li2020attention,dai2020spsg,dai2020sg,hou2020revealnet,najibi2020dops,dai2018scancomplete}. This strategy also enables decoding voxel labels to complete scenes at the semantic or instance level. However, the expensive 3D convolutions in the scene level make them suffer from the resolution problem. Shape retrieval \cite{avetisyan2019scan2cad,avetisyan2019end,avetisyan2020scenecad,ishimtsev2020cad} provides an alternative method to predict shapes by searching for a CAD model as similar to the incomplete object as possible. However, the accuracy and the computation efficiency depends on the model dataset scale. Compared to voxels/TSDFs, point clouds present sparsity and are more scalable for efficient learning. To our knowledge, few works have attempted to learn object meshes at the semantic-instance level directly from points.

In this work, we provide a \textit{reconstruction-from-detection} framework, \textit{RfD-Net}, for end-to-end semantic instance reconstruction directly from raw point clouds (see Figure~\ref{fig:teaser}). Our design is based on the insight that object detections provide spatial alignment that enables better local shape reconstruction. On the other hand, object shapes in sparse point clouds indicate local geometry that should back improve 3D detection. It decouples the problem of scene reconstruction into global localization and local shape prediction. Our method embeds the shape reconstruction head with a 3D detector backbone. It leverages the sparsity of point clouds and focuses on predicting shapes that are detected with high objectness and ignores the free scene space. With this design, our method allows implicit function learning to reconstruct surfaces with much higher resolution. In our experiments, we observe that joint shape reconstruction and 3D detection presents complementary effects. Deploying the shape prediction head is shown consistently effective on improving the modern point-based detectors, and vice versa, which makes our method consistently outperforms the prior arts in 3D detection and instance reconstruction. In summary, we list our contributions as follows:
\begin{itemize}[leftmargin=*]
	\item We provide a novel learning modality for semantic instance reconstruction. To our knowledge, it is the first learning method to predict instance semantics with geometry directly from point clouds. While prior methods heavily rely on 3D CNNs to learn from voxelized scenes.
	\item We propose a new end-to-end architecture, namely \textit{RfD-Net}, to learn object semantics and shapes from sparse point clouds. It disentangles semantic instance reconstruction into global object localization and local shape prediction, which are bridged with a skip propagation module to facilitate joint learning.  With this manner, our shape generator supports implicit learning, which directly overcomes the resolution bottleneck in the prior art \cite{hou2020revealnet}.
	\item Joint learning object poses and shapes presents complementary benefits. It also shows consistent effects on modern detection backbones, and makes our method achieve the state-of-the-art in instance detection\&completion and improve over 11 of mesh IoU in object reconstruction.
\end{itemize}

\section{Related Work}
In this section, we mainly summarize the recent works of 3D deep learning in shape completion, scene completion and instance-level scene reconstruction from point clouds.

\noindent\textbf{Shape Completion.} Shape completion aims to recover the missing geometries of a target object from a partial scan. From point cloud inputs, many methods complete shapes represented by points \cite{yuan2018pcn,tchapmi2019topnet,liu2020morphing,nie2020skeleton,huang2020pf,wang2020cascaded}, voxels \cite{dai2017shape,han2017high}, SDFs, \cite{liao2018deep,mescheder2019occupancy,chibane2020implicit} and meshes \cite{groueix2018papier,hanocka2020point2mesh}. Most of them share the similar modality, i.e., to encode the incomplete scan with point-wise convolutions (e.g., PointNet or PointNet++ \cite{qi2017pointnet,qi2017pointnet++}) and predict the invisible parts, holes or defective surfaces, while preserving the shape topology. They more focus on the completeness of single targets. In this work, we take advantage of the recent advance in shape completion for scene reconstruction. It supports our method to reconstruct high-resolution objects in a 3D scene.

\noindent\textbf{Scene Completion.} Scene completion focuses on predicting all objects including visible and invisible geometries from an incomplete scan. Different from shape completion, occlusions between objects manifest the major challenge. To this end, some works attempt to inpaint depth frames in scanning to recover object surfaces \cite{han2019deep,novotny2019perspectivenet,huang2019indoor,senushkin2020decoder}. Benefited from 2D CNNs, these methods achieve impressive results in surface points recovery. Similarly, more works extend the advantage of CNNs to 3D \cite{firman2016structured,dai2020spsg,dai2020sg}. They discretize incomplete scans into voxel or TSDF grids and predict per-voxel occupancy with 3D CNNs for scene completion. With the fully convolution design, it also supports scene understanding tasks joint with completion, such as semantic completion or segmentation \cite{song2017semantic,zhang2018efficient,li2019depth,liu2018see,wang2019forknet,wang2020deep,li2020anisotropic,li2020attention,chen20203d,hou2020revealnet,dai2018scancomplete}. However, the expensive 3D CNNs consume much computation that hinders them with limited resolution, which appears more obvious on instance objects. In our work, we directly learn from point clouds. Objects detected with high objectness are only considered, which enables us to predict meshes with much higher resolution.

\begin{figure*}[!t]
	\centering
	\includegraphics[width=\textwidth]{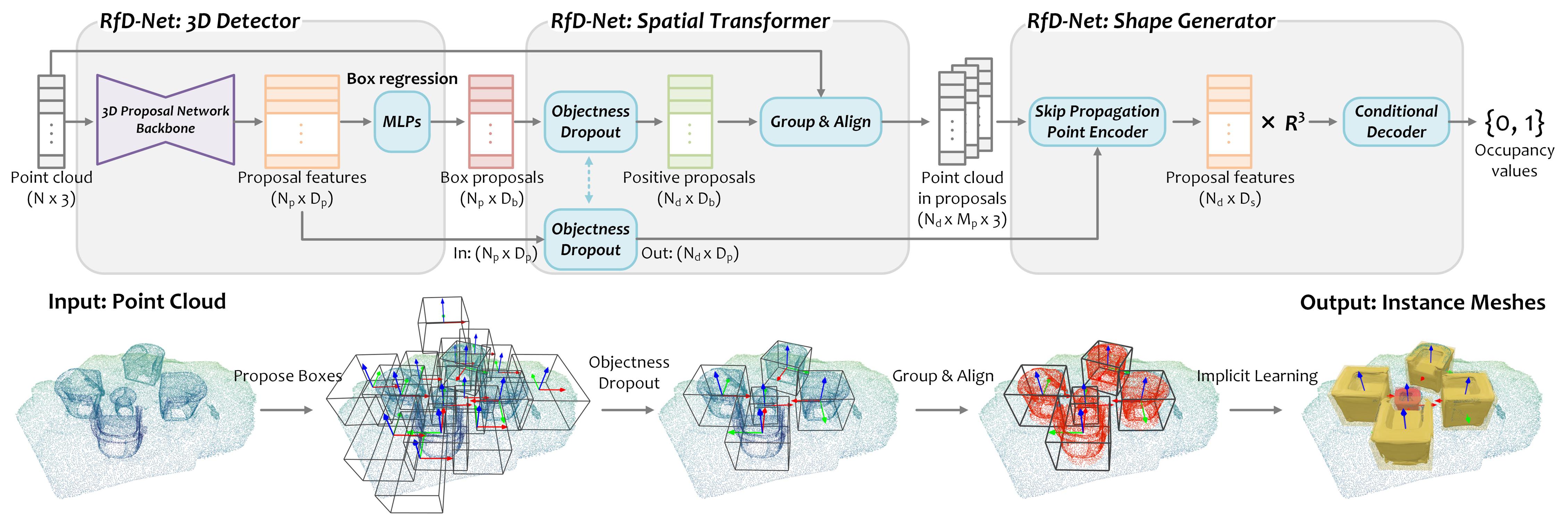}
	\caption{Overview of the network architecture. From an input point cloud with $N$ points, a 3D proposal network backbone proposes $N_{p}$ object proposals ($D_{p}$-dim features) that are decoded into $D_{b}$-dim box parameters. With a Top-$N$ rank dropout layer, we reserve $N_{d}$ proposals with higher objectness in the sparse point cloud. This subset of proposals are considered to independently group \& align the neighboring $M_{p}$ point cloud into clusters. Each point cluster is encoded into a $D_{s}$-dim vector through our point encoder to regress the binary occupancies of spatial query points ($x$-$y$-$z$) for mesh generation.}
	\label{fig:overview}
	\vspace{-15pt}
\end{figure*}

\noindent\textbf{Instance Reconstruction.} Beyond recovering scene geometries. Instance reconstruction refers to both object localization and reconstruction. With the advent of 3D scanners, early works \cite{nan2012search,shao2012interactive,kim2012acquiring,chen2014automatic} focus on an approximate solution, i.e. semantic modeling. Object shapes are retrieved with CAD models or primitives with non-linear optimization. After that, 3D deep learning reforms this process into a learnable manner that model retrieval can be replaced with deep feature matching \cite{avetisyan2019scan2cad,avetisyan2019end,avetisyan2020scenecad,ishimtsev2020cad}. Semantic modeling presents delicate shape models though, the matching similarity and inference efficiency directly rely on the CAD dataset scale. The closest topic to us is semantic instance reconstruction (or completion). \cite{najibi2020dops} explores the possibility of predicting shapes along with object detection from voxelized point clouds. The previous state-of-the-art \cite{hou2020revealnet} discretizes the 3D scan into TSDF grids and predicts the occupancy of semantic instances with 3D CNNs. As we mentioned, their scene resolution ($\approx$96$\times$48$\times$96) is limited by the heavy 3D CNN computation. Differently, we leverage the sparsity of point clouds where objects detected with high objectness will be reconstructed. It supports single object reconstruction to recover high-resolution shapes and jointly improves 3D object detection performance.

\section{Method}
We illustrate the architecture of \textit{RfD-Net} in Figure~\ref{fig:overview}. Our method follows the basic principle of understanding 3D scenes with `\textit{reconstruction from detection}'. On this top, we devise the network consisting of a \textit{3D detector}, a \textit{spatial transformer} and a \textit{shape generator}. We build this architecture as generic as possible for learning instance shapes from point clouds, which should be flexibly compatible to modern point-based 3D proposal network backbones \cite{qi2019deep,xie2020mlcvnet}. Specifically, from an input point cloud, the 3D detector generates box proposals to locate object candidates from the sparse 3D scene. Then, we design a spatial transformer to select positive box proposals and \textit{group \& align} the local point cloud within for the next object shape generation. Shape generator independently learns an occupancy function in a canonical system to represent the shape of proposals. We elaborate on the details of each module as follows.

\subsection{Learning Object Proposals in Point Clouds}
\label{sec:detection}
From the input point cloud $\mathbf{P}\in\mathbb{R}^{N\times 3}$, we adopt the VoteNet \cite{qi2019deep} as the backbone to propose $N_{p}$ box proposals with $D_{p}$-dim features $\bm{F}_{p}\in\mathbb{R}^{N_{p}\times D_{p}}$. We use $\bm{F}_{p}$ to predict $D_{b}$-dim box parameters, including center $\bm{c}\in\mathbb{R}^{3}$, scale $\bm{s}\in\mathbb{R}^{3}$, heading angle $\theta\in\mathbb{R}$, semantic label $l$ and objectness score $s_{obj}$ (as parameterized in \cite{qi2019deep}). The objectness score is to classify if a proposal is close to (\textless0.3 m, positive) or far from (\textgreater0.6 m, negative) any ground-truth object center. We regress the box parameters with a 2-layer MLP.

In this part, each proposal feature $\bm{f}_{p}$ in $\bm{F}_{p}$ summarizes the semantics and box geometry information of a local region, which is of higher-level compared to the raw input points. In our method, $\bm{F}_{p}$ are also used to extend the local points for shape generation with our skip propagation module (see Section~\ref{sec:shapegeneration}). It enhances the gradient back-propagation from shape generation to object detection, and is shown consistently effective on improving the object detection performance with different 3D detection backbones (see analysis experiments in section~\ref{sec:analysis_experiments}).

\begin{figure*}
	\centering
	\includegraphics[width=1\linewidth]{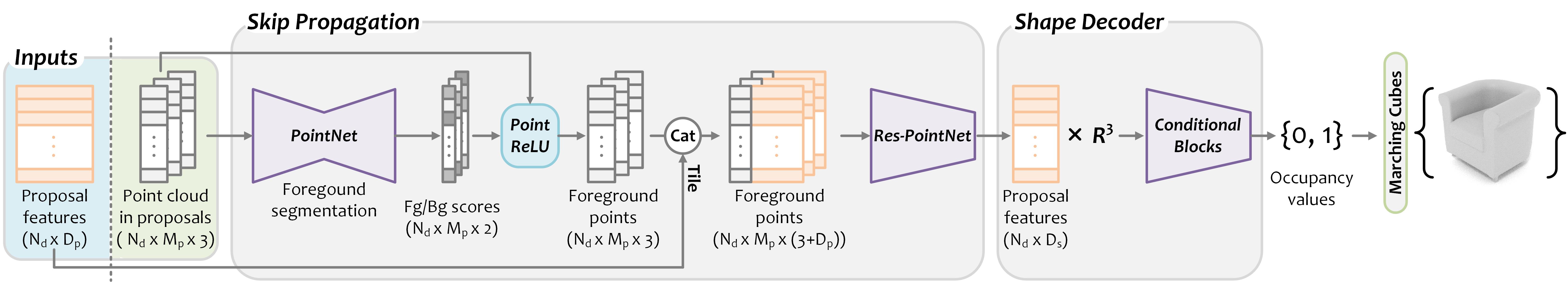}
	\caption{Shape generation from object proposals. From the 3D points in the positive $N_{d}$ object proposals, our method first learns a denoiser (with PointNet \cite{qi2017pointnet} layers) to remove the background points. These cleaned points are afterwards extended by the former proposal box features with a skip connection, before being encoded into new proposal features to decode spatial occupancy values. The object meshes are obtained by Marching Cubes \cite{lorensen1987marching} on the predicted occupancy grids.}
	\label{fig:shapegen}
	\vspace{-15pt}
\end{figure*}

\subsection{Transforming Points to Local}
\label{sec:transformer}
\noindent\textbf{Objectness Dropout.}
To obtain the object points for shape generation, we develop a spatial transformer to group the input point cloud in box proposals, where object points in each proposal are transformed and aligned into a local canonical system (see Figure~\ref{fig:overview}). Before the group \& align process, we only reserve $N_{d}$ positive boxes out of all the proposals during training. We adopt Top-$N$ rank dropout to reserve $N_{d}$ proposals with higher objectness ($N_{d}=10$). Different from images, 3D point cloud presents severe sparsity that free space occupies the major area. Too many negative samples (e.g., boxes with no/few points or far from any object) will be involved if considering all box proposals, which prevents the shape generator from learning correct shapes. Besides, the large computation also undermines the method efficiency. The effects with different numbers of positive proposals are discussed in our experiments. During inference, we replace the objectness dropout layer with a 3D NMS module to produce the output boxes.

\noindent\textbf{Group \& Align.}
Within the $N_{d}$ positive box proposals, we sample points from the input point cloud. We group points that are located within a radius $r$ to these box centers $\left\{\bm{c}_{i}\right\}$ using a group layer \cite{qi2017pointnet++}. It produces $N_{d}$ clusters and $M_{p}$ in-cluster points for each, where we denote the point clusters by $\left\{\mathbf{P}^{c}_{i}\right\}$ ($i=1,2,...,N_{d}, \mathbf{P}^{c}_{i}\in\mathbb{R}^{M_{p}\times3}$). Afterwards, we align the points in different clusters into a canonical coordinate system. It is to normalize the input points, which removes the variance in spatial translation and rotation for shape prediction. We formulate this process by:
\begin{equation}
	\label{eqn:01}
	\begin{aligned}
		\tilde{\mathbf{P}}^{c}_{i} &=\left[\mathcal{R}^{-1}\left(\theta_{i}\right)+\Delta\mathcal{R}\right]\cdot\left[\mathbf{P}^{c}_{i}-\left(\bm{c}_{i}+\Delta\bm{c}\right)\right],
	\end{aligned}
\end{equation}
where $\tilde{\mathbf{P}}^{c}_{i}$ denotes the aligned point cluster. $\bm{c}_{i}$ and $\theta_{i}$ represents the corresponding box center and heading angle predicted from the 3D detector. $\mathcal{R}\left(\cdot\right)\in\mathbb{R}^{3\times3}$ is the rotation matrix. Since there would be deviations between the predicted centers \& heading angles and the corresponding ground-truths, we learn an adjustment $\left(\Delta\mathcal{R},\Delta\bm{c}\right)$ from the input point cluster $\mathbf{P}^{c}_{i}$ by:
\begin{equation}
	\label{eqn:02}
	\begin{aligned}
		\left[\Delta\mathcal{R},\Delta\bm{c}\right] &=\text{MLP}_{2}\left\{\max_{\bm{p} \in \mathbf{P}^{c}_{i}}\left\{\text{MLP}_{1}\left(\bm{p}\right)\right\}\right\}.
	\end{aligned}
\end{equation}
In Equation~\ref{eqn:02}, 3D points in each cluster $\mathbf{P}^{c}_{i}$ are independently processed with $\text{MLP}_{1}$, and max-pooled point-wisely into a global feature vector before being regressed into $\left(\Delta\mathcal{R},\Delta\bm{c}\right)$ by $\text{MLP}_{2}$. So far, we have obtained the aligned 3D points $\{\tilde{\mathbf{P}}^{c}_{i}\}$ with the corresponding proposal features $\left\{\bm{f}_{p}\right\}\subset\bm{F}_{p}$ of the positive box proposals.

\subsection{Shape Generation from Proposals}
\label{sec:shapegeneration}
We illustrate the shape generation module in Figure~\ref{fig:shapegen}. Our shape generator consists of two parts: a skip propagation encoder to extend proposal features $\left\{\bm{f}_{p}\right\}$ with cluster points $\{\tilde{\mathbf{P}}^{c}_{i}\}$; a shape decoder to learn the occupancy of spatial points conditioned on the extended proposal features.
 
\noindent\textbf{Skip Propagation.}
From the 3D points in proposals $\{\tilde{\mathbf{P}}^{c}_{i}\}$, we first deploy a denoiser to reserve the foreground points for shape generation. It is implemented with PointNet \cite{qi2017pointnet} layers to classify if a point belongs to the foreground. We remove these background points with a point-wise ReLU layer. Foreground points in each proposal are afterward concatenated with the corresponding proposal feature $\bm{f}_{p}$ from the 3D detector by a skip connection. We propagate the upstream proposal feature to each 3D point, since it has summarized the information of object category and 3D size. We use it as shape priors to inform the decoder for efficient shape approximation. The skip propagation module bridges the gradient back-propagation from the shape generator to 3D detector and jointly improves shape prediction with object detection (see ablations in section~\ref{sec:comp_to_sota}). From the extended point clusters, we further deploy a PointNet \cite{qi2017pointnet} designed with residual connections to encode them into a set of new proposal features $\left\{\bm{f}^{*}_{p}\in\mathbb{R}^{D_{s}}\right\}$.

\begin{figure}
	\centering
	\includegraphics[width=0.8\linewidth]{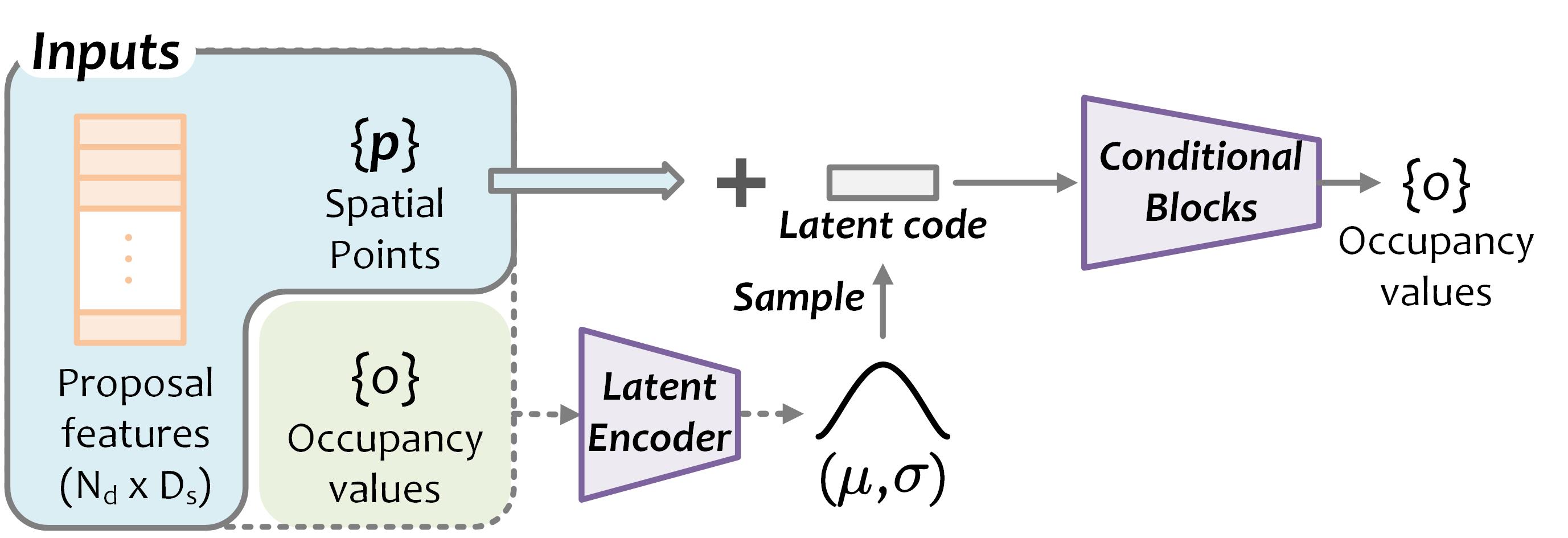}
	\caption{Probabilistic shape generation in training.}
	\label{fig:latentdecoder}
	\vspace{-15pt}
\end{figure}

\noindent\textbf{Shape Decoder.}
We implicitly represent a 3D shape as an occupancy function \cite{mescheder2019occupancy,chibane2020implicit,saito2019pifu}. That is, from spatial points $\left\{\bm{p}\in\mathbb{R}^{3}\right\}$ sampled in the shape bounding cube, we learn to predict their binary occupancy value $o\in\left\{0,1\right\}$ (inside/ouside the shape) conditioned on the input observation ($\bm{f}^{*}_{p}$). As in \cite{mescheder2019occupancy}, we adopt the Conditional Batch Normalization \cite{de2017modulating,dumoulin2016adversarially} layers to make the decoder conditioned on the proposal features to regress occupancy values. Furthermore, since the object point clouds are usually partially scanned, they could have multiple shape explanations. To this end, we build this decoder as a probabilistic generative model \cite{wu2016learning,achlioptas2018learning} (see Figure~\ref{fig:latentdecoder}). Specifically, from the input points $\left\{\bm{p}\right\}$, occupancies $\left\{o\right\}$ and the proposal feature $\bm{f}^{*}_{p}$, we adopt the latent encoder in \cite{mescheder2019occupancy} and predict the mean and standard deviation $\left(\bm{\mu}, \bm{\sigma}\right)$ to approximate the standard normal distribution. We sample on our distribution to produce a latent code $z\in\mathbb{R}^{L}$. $\left\{\bm{p}\right\}$ and $z$ are then processed into equal dimension with a single layer of MLP before summed point-wisely, and fed to five conditional blocks to regress the occupancies $\left\{o\right\}$. During inference, the latent code $z$ is set to zeros, and we use the Marching Cubes algorithm \cite{lorensen1987marching} to extract the mesh surface from spatial occupancy grids.

\section{End-to-end Learning}
In this section, we summarize the learning targets with corresponding losses for end-to-end training.

\noindent\textbf{Box Loss.} The 3D detector predicts the proposal objectness score $s_{obj}$, box center $\bm{c}$, scale $\bm{s}$, heading angle $\theta$ with the semantic label $l$. As in \cite{qi2019deep}, we supervise the objectness loss ($\mathcal{L}_{obj}$) of box proposals with cross entropy, to classify if they are located within 0.3 meters (positive) or more than 0.6 meters (negative) to any object center. Proposals with positive objectness are supervised with the following box loss. We supervise the box center loss $\mathcal{L}_{\bm{c}}$ using Smooth-$L_{1}$ loss, and follow \cite{qi2018frustum,huang2018cooperative} to disentangle the scale loss $\mathcal{L}_{\bm{s}}$ and the heading angle loss $\mathcal{L}_{\theta}$ into a hybrid of classification (cross entropy) and regression (Smooth-$L_{1}$) losses, i.e., $ \lambda_{cls}\mathcal{L}_{cls} + \mathcal{L}_{reg}$. We use the cross-entropy as the semantic classification loss ($\mathcal{L}_{l}$). Each proposal is paired with the nearest ground-truth box for training. For using VoteNet as the backbone, we involve an extra vote loss $\mathcal{L}_{v}$ as in \cite{qi2019deep}. So far the above losses can be concluded into the box loss $\mathcal{L}_{box}$ with weighted sum.

\noindent\textbf{Shape Loss.} For points in each proposal, we supervise the foreground segmentation loss $\mathcal{L}_{seg}$ with cross entropy. The shape generator learns $\left(\bm{\mu}, \bm{\sigma}\right)$ to approximate the standard normal distribution in training, from which we sample a latent code $\bm{z}$ to predict the occupancy values $\left\{o\right\}$ of spatial query points $\left\{\bm{p}\right\}$ with a condition of the proposal feature $\bm{f}^{*}_{p}$. Then the $\mathcal{L}_{shape}$ can be formulated by
\begin{equation}
	\label{eqn:04}
	\begin{aligned}
		\mathcal{L}_{shape}
		=&\frac{1}{N_{d}}\sum_{i=1}^{N_{d}}\Big[\sum_{j=1}^{K}\mathcal{L}_{ce}\left(\hat{o}_{i,j}, o_{i,j}\right) \\ &+ \text{KL}\left(\hat{p}\left(\bm{z}_{i}\right)\| p\left(\bm{z}_{i}\right)\right)\Big] + \lambda_{seg}\mathcal{L}_{seg},
	\end{aligned}
\end{equation}
where $\mathcal{L}_{ce}$ and KL denotes the cross entropy loss and KL-divergence respectively. $\hat{o}_{i,j}$ and $o_{i,j}$ correspond to the predicted and ground-truth occupancy of the $j$-th spatial point in the $i$-th positive proposal. $\hat{p}\left(\bm{z}_{i}\right)$ is the predicted distribution of the latent code $\bm{z}_{i}$. $p\left(\bm{z}_{i}\right)$ is the target distribution, which is set to the standard normal distribution as \cite{mescheder2019occupancy}.

Overall, we train our network end-to-end with the loss of $\mathcal{L}=\mathcal{L}_{box} + \lambda\mathcal{L}_{shape}$. The weights to balance different losses are detailed in the supplementary material.

\section{Results and Evaluation}
\subsection{Experiment Setup}
\noindent\textbf{Data.}
Two datasets are used in our experiments. 1) ScanNet v2 \cite{dai2017scannet} consists of 1,513 real world scans with point clouds labeled at the instance level. 2) Scan2CAD \cite{avetisyan2019scan2cad} aligns the ShapeNet \cite{chang2015shapenet} models with the object instances in ScanNet. It provides the object meshes. We preprocess these object meshes following \cite{mescheder2019occupancy} to prepare the spatial points and occupancy values for shape learning. We only use the point clouds from ScanNet as the input (randomly sampled with 80K points in train and test), and predict object bounding boxes and meshes supervised by Scan2CAD. Inline with \cite{hou2020revealnet}, we use the official train/test split in all the experiments and consider eight object categories.

\noindent\textbf{Metrics.}
We evaluate our method on both scene understanding and object reconstruction, including 3D detection, single object reconstruction, and semantic instance completion. As \cite{hou2020revealnet}, the mean average precision at the 3D IoU threshold of 0.5 (mAP@0.5) is used in the evaluation of 3D detection and semantic instance completion, and we adopt 3D IoU for mesh evaluation in single object reconstruction.

\noindent\textbf{Implementation.}
Our network is end-to-end trained before pretraining the 3D proposal network. We set the batch size at 8, and adopt the initial learning rate at 1e-3 in pretraining and 1e-4 in end-to-end training, which drops by the scale of 0.5 for every 80 epochs. 240 epochs are used in total. During inference, we replace the objectness dropout (see Section~\ref{sec:detection}) with 3D NMS \cite{qi2019deep} to output the 3D boxes, where object meshes in these boxes are extracted from 128\textsuperscript{3} occupancy grids and reconstructed with Marching Cubes \cite{lorensen1987marching}. All object meshes are predicted in a canonical system and transformed to 3D scenes with the predicted box poses. We provide the full list of parameters, layer specifications, and efficiency\&memory comparisons in the supplementary file.

\noindent\textbf{Benchmark.}
We compare our method on the metrics above with the previous state-of-the-art works in scene reconstruction, including RevealNet \cite{hou2020revealnet}, 3D-SIS \cite{hou20193d} and ScanComplete \cite{dai2018scancomplete}. We also investigate the compatibility of our method by embedding with different 3D proposal network backbones, including BoxNet \cite{qi2019deep}, MLCVNet \cite{xie2020mlcvnet}, and VoteNet with Graph Voting \cite{dgcnn}. All results in our experiments are trained and tested with the same dataset and split. We keep the geometric data of 3D scans as the input in comparisons. For some methods requiring color images, we keep inline with their original configuration.

\begin{figure*}[!ht]
	\centering
		\begin{subfigure}[t]{0.24\textwidth}
		\includegraphics[width=\textwidth]  
		{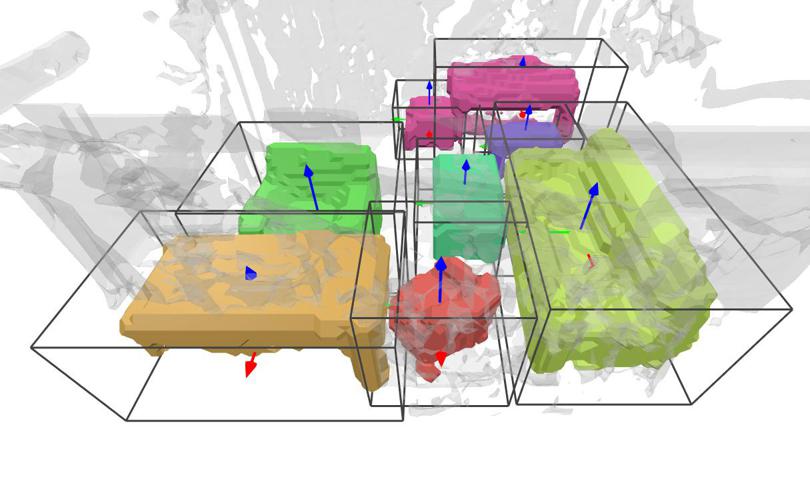}
		\includegraphics[width=\textwidth]
		{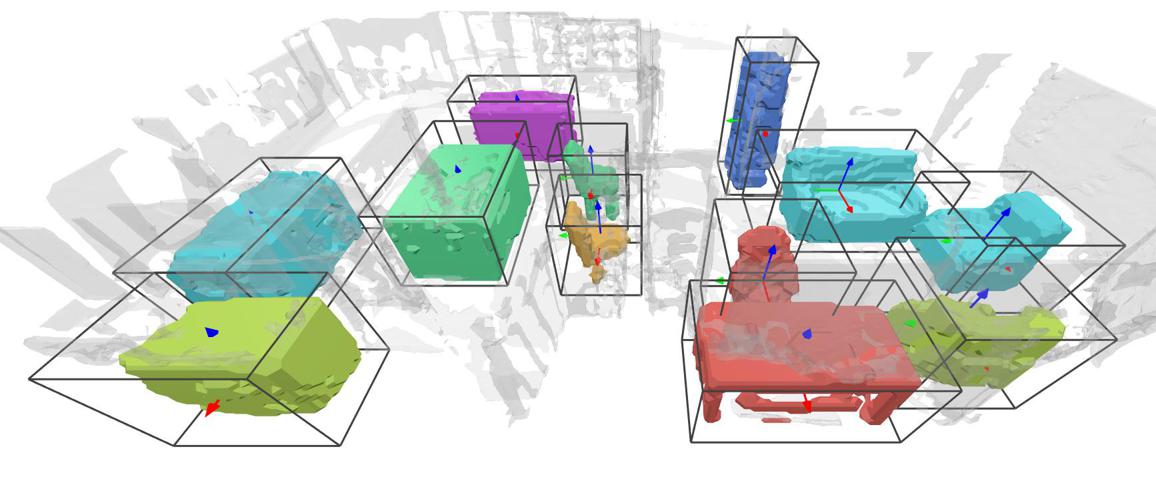}
		\includegraphics[width=\textwidth]
		{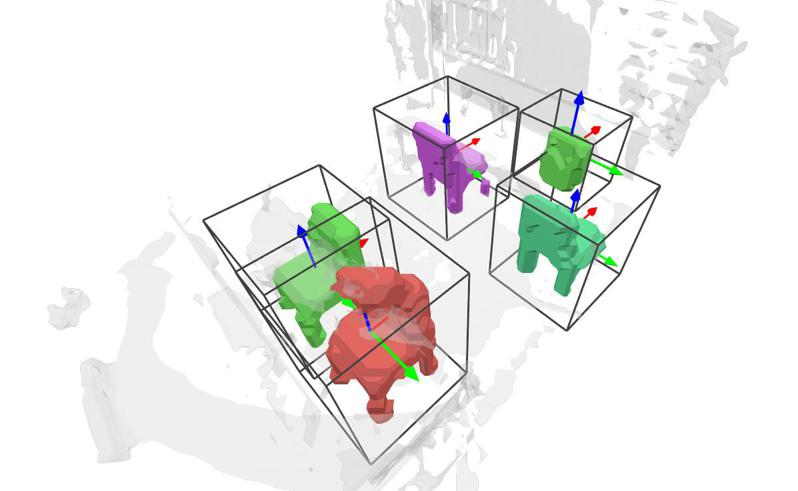}
		\includegraphics[width=\textwidth]
		{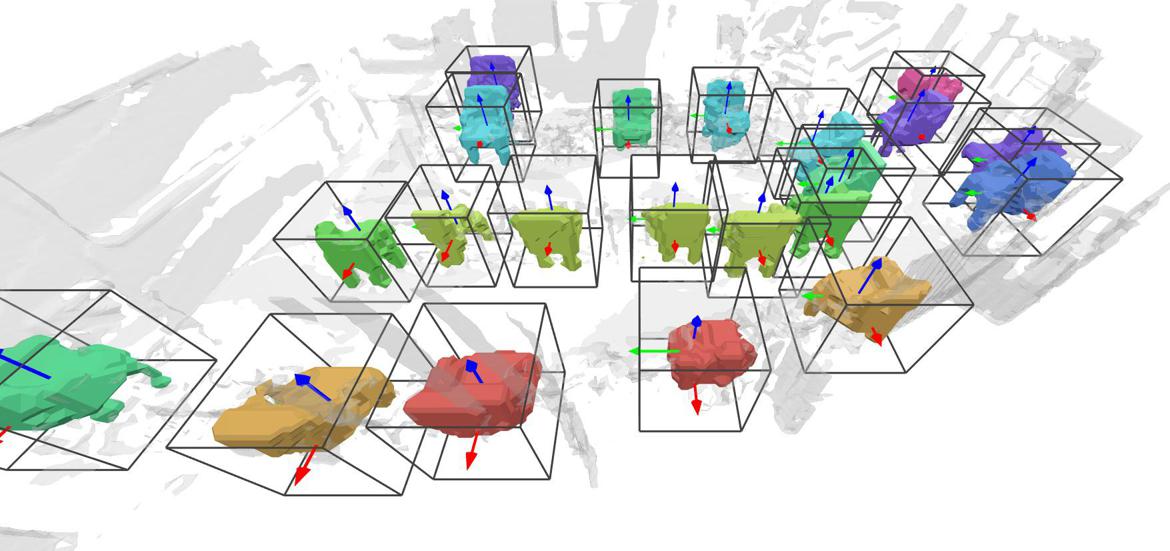}
		\includegraphics[width=\textwidth]
		{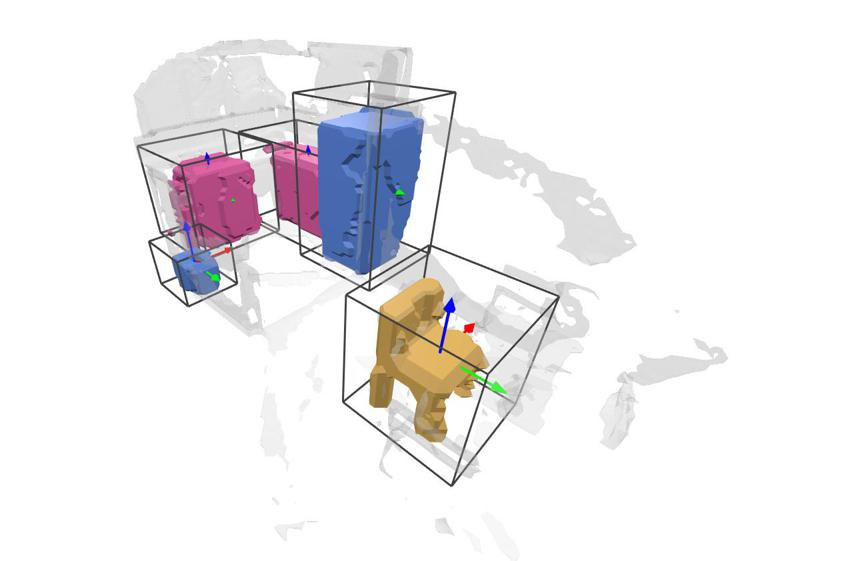}
		\includegraphics[width=\textwidth]
		{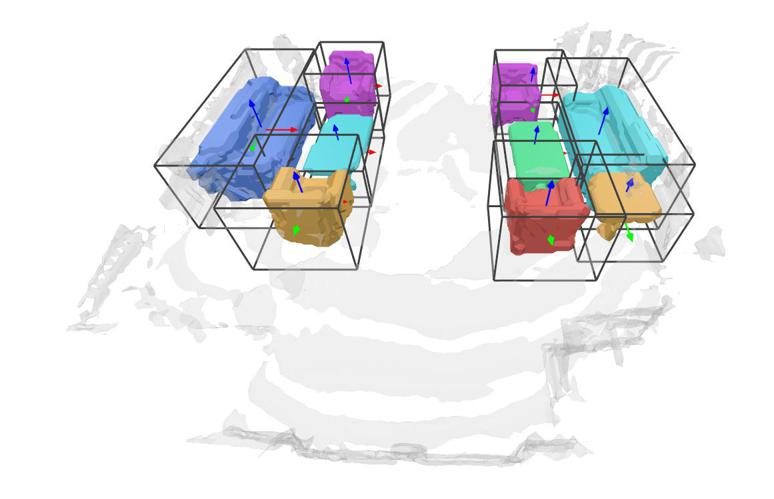}
		\caption{RevealNet \cite{hou2020revealnet} (Geo+Image)}
	\end{subfigure}
	\begin{subfigure}[t]{0.24\textwidth}
		\includegraphics[width=\textwidth]
		{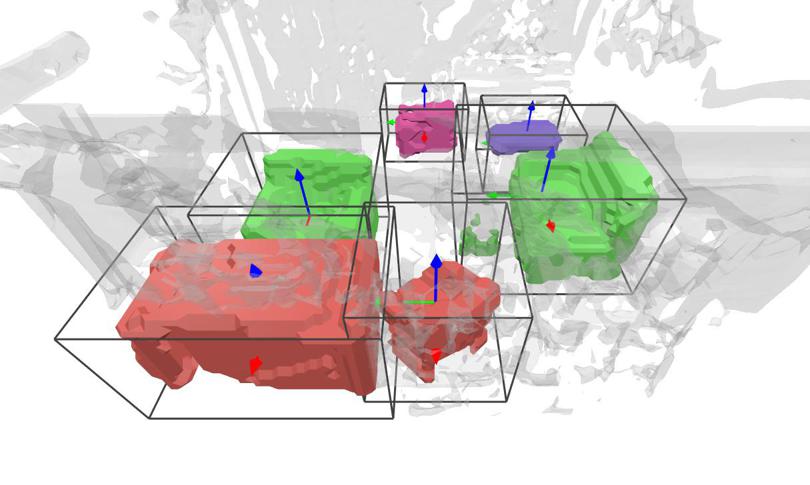}
		\includegraphics[width=\textwidth]
		{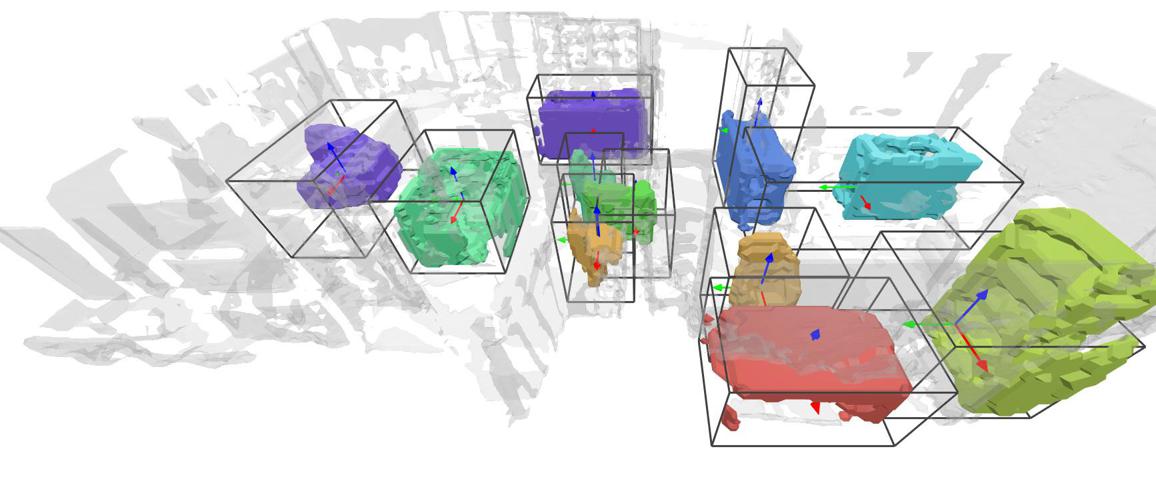}
		\includegraphics[width=\textwidth]
		{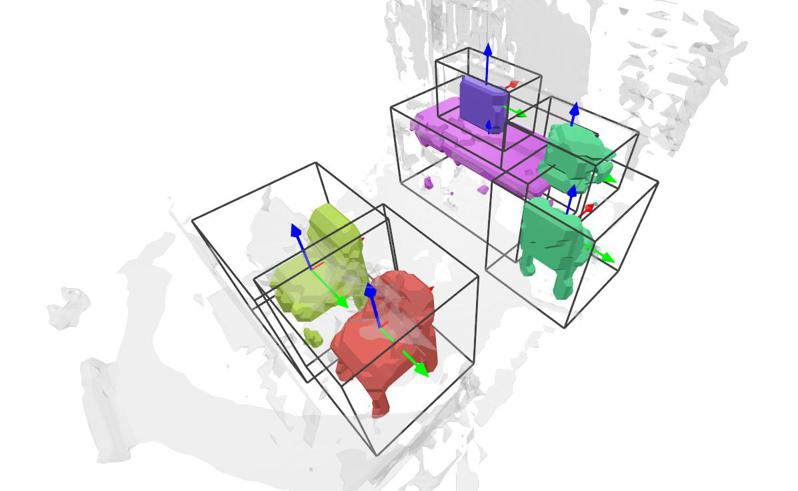}
		\includegraphics[width=\textwidth]
		{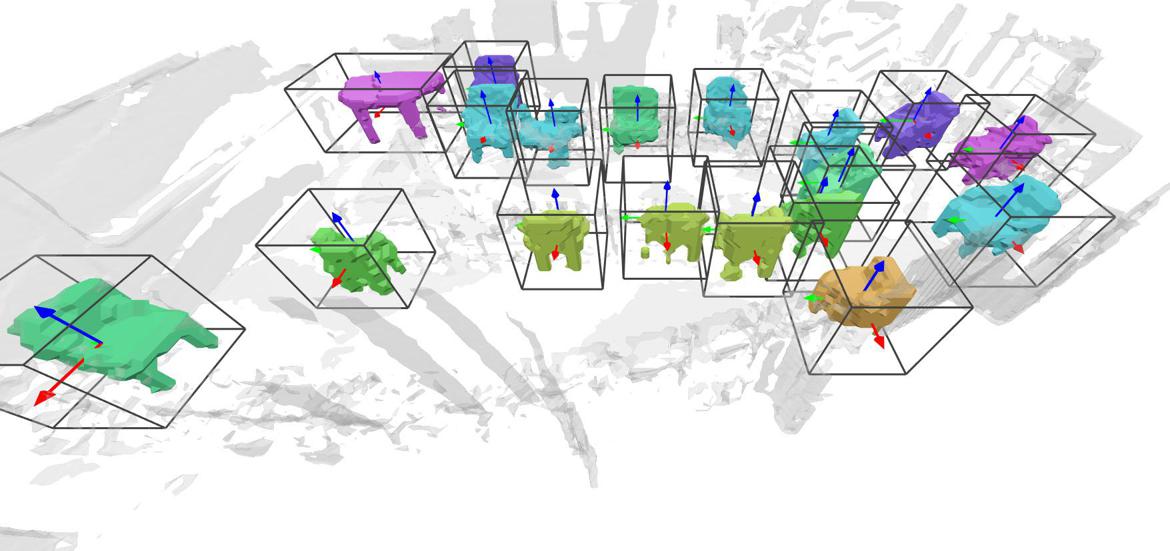}
		\includegraphics[width=\textwidth]
		{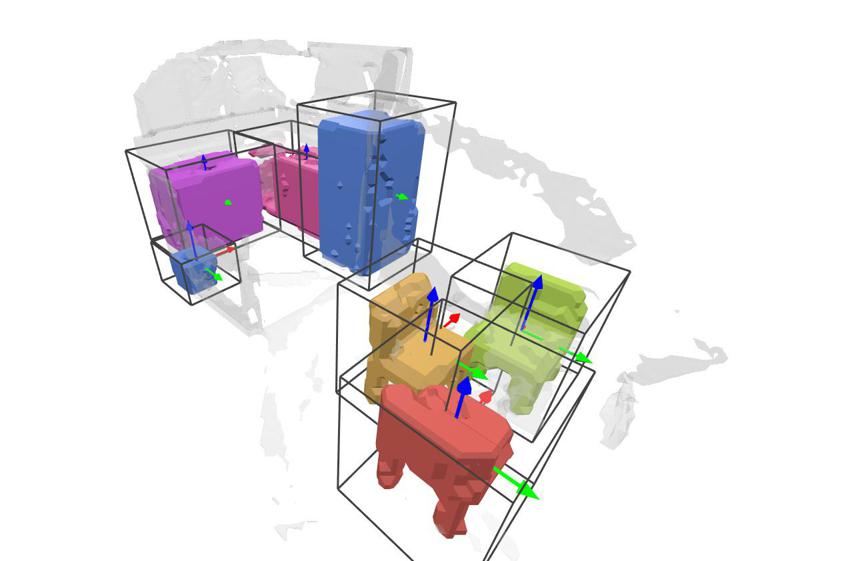}
		\includegraphics[width=\textwidth]
		{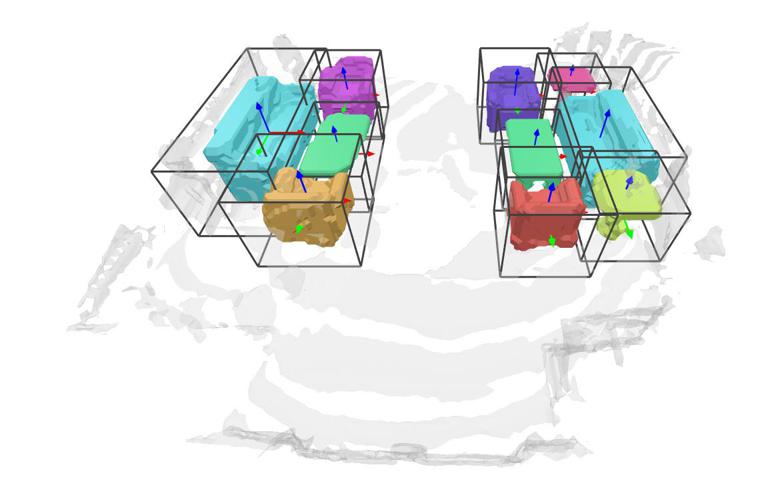}
		\caption{RevealNet \cite{hou2020revealnet} (Geo Only)}
	\end{subfigure}
	\begin{subfigure}[t]{0.24\textwidth}
		\includegraphics[width=\textwidth]  
		{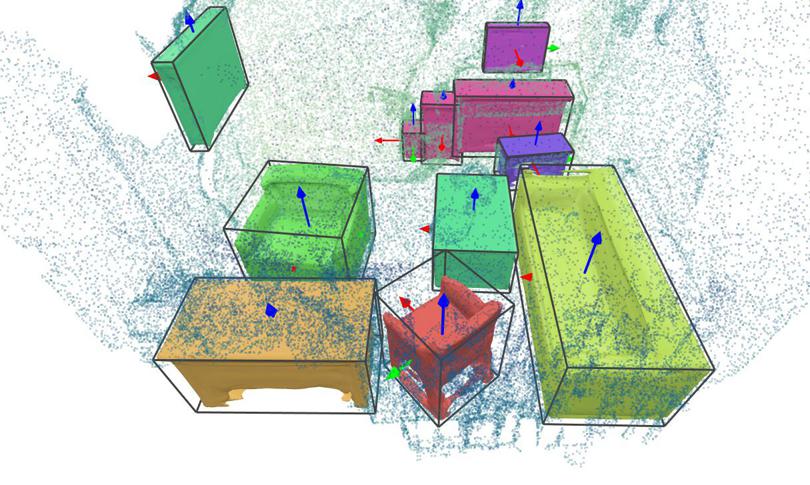}
		\includegraphics[width=\textwidth]
		{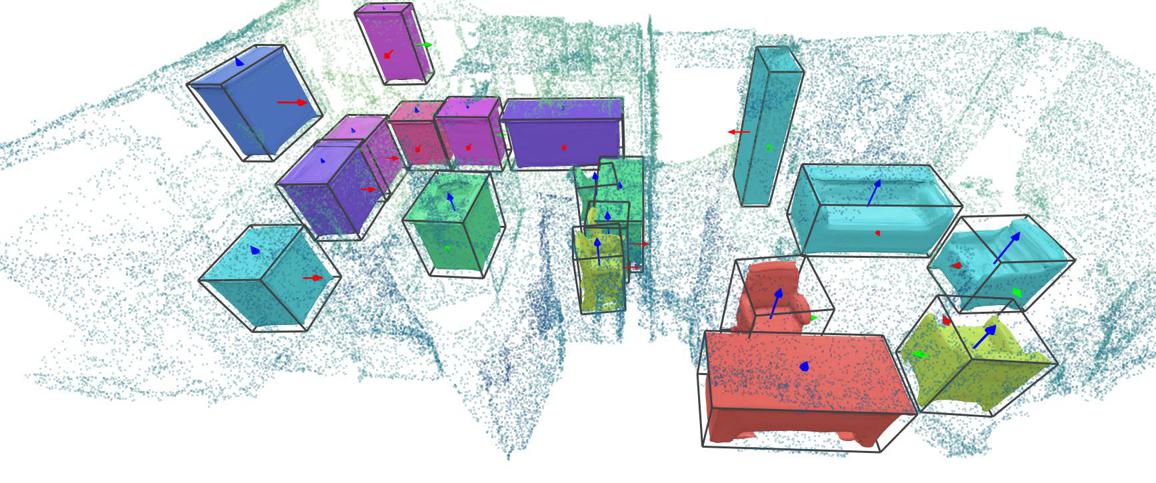}
		\includegraphics[width=\textwidth]
		{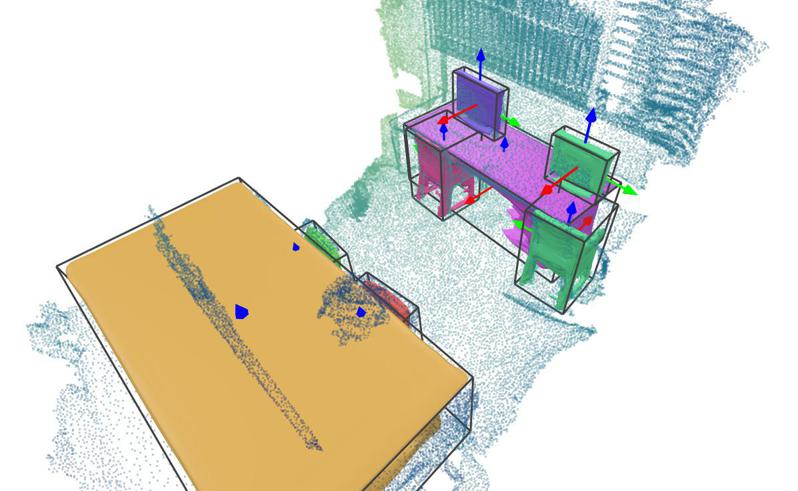}
		\includegraphics[width=\textwidth]
		{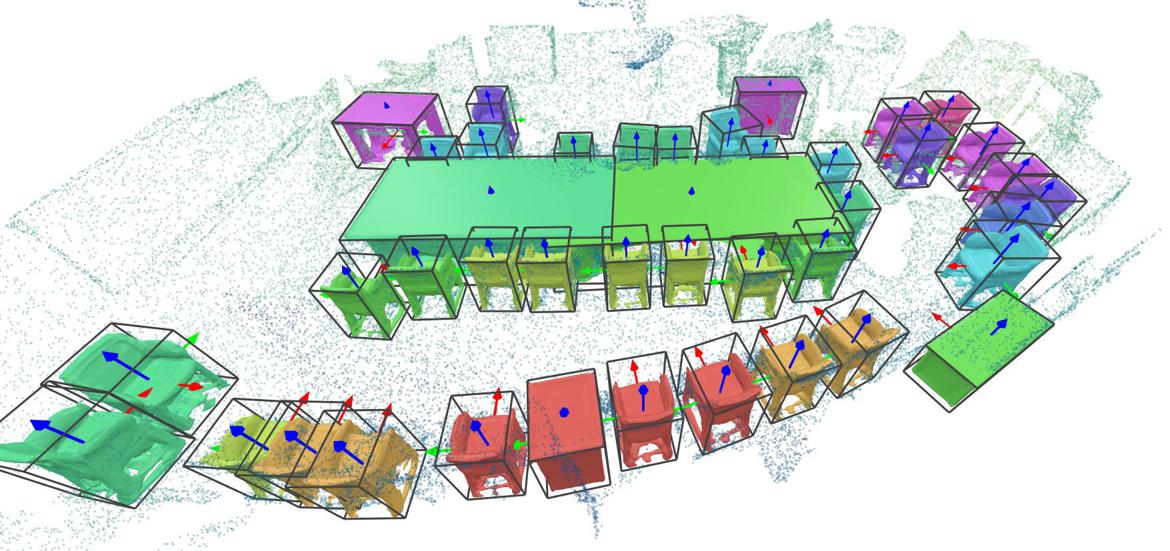}
		\includegraphics[width=\textwidth]
		{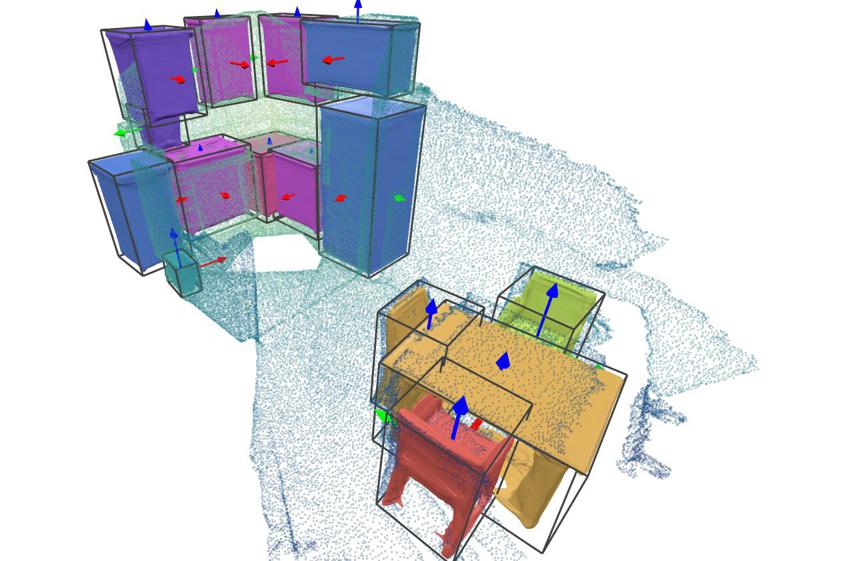}
		\includegraphics[width=\textwidth]
		{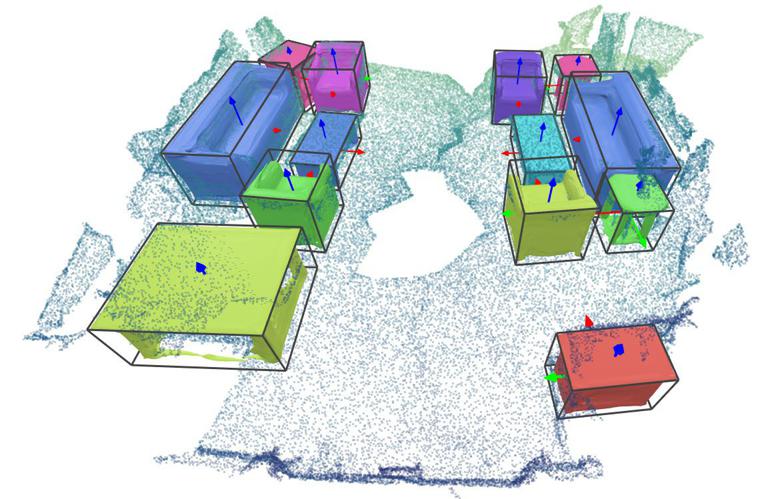}
		\caption{Ours (Geo Only)}
	\end{subfigure}
	\begin{subfigure}[t]{0.24\textwidth}
		\includegraphics[width=\textwidth]  
		{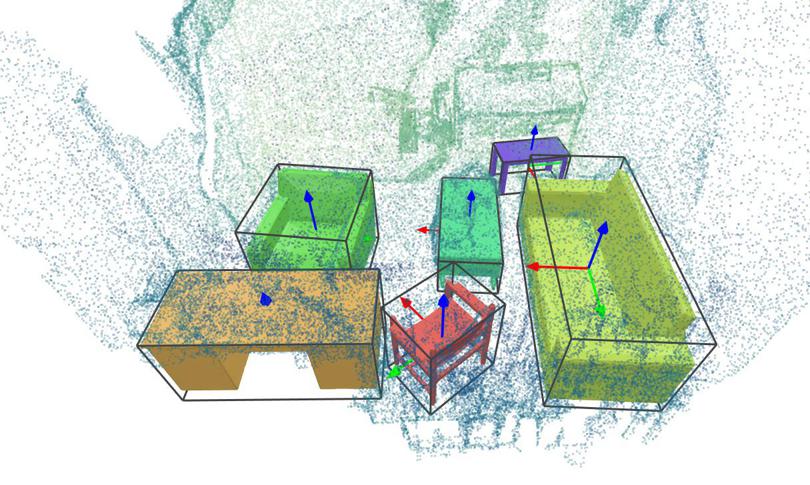}
		\includegraphics[width=\textwidth]
		{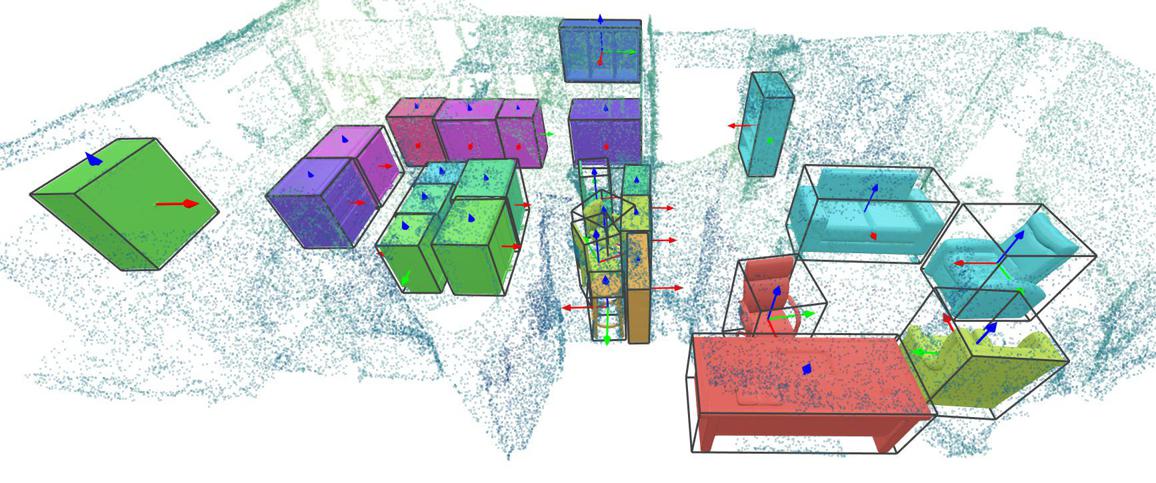}
		\includegraphics[width=\textwidth]
		{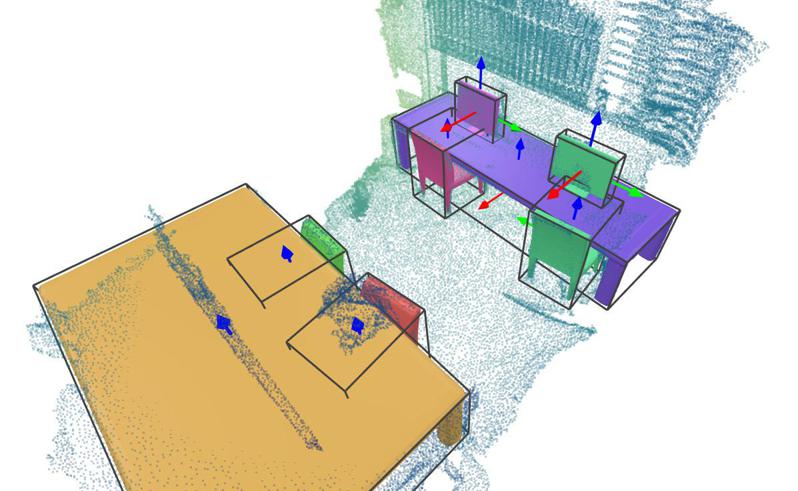}
		\includegraphics[width=\textwidth]
		{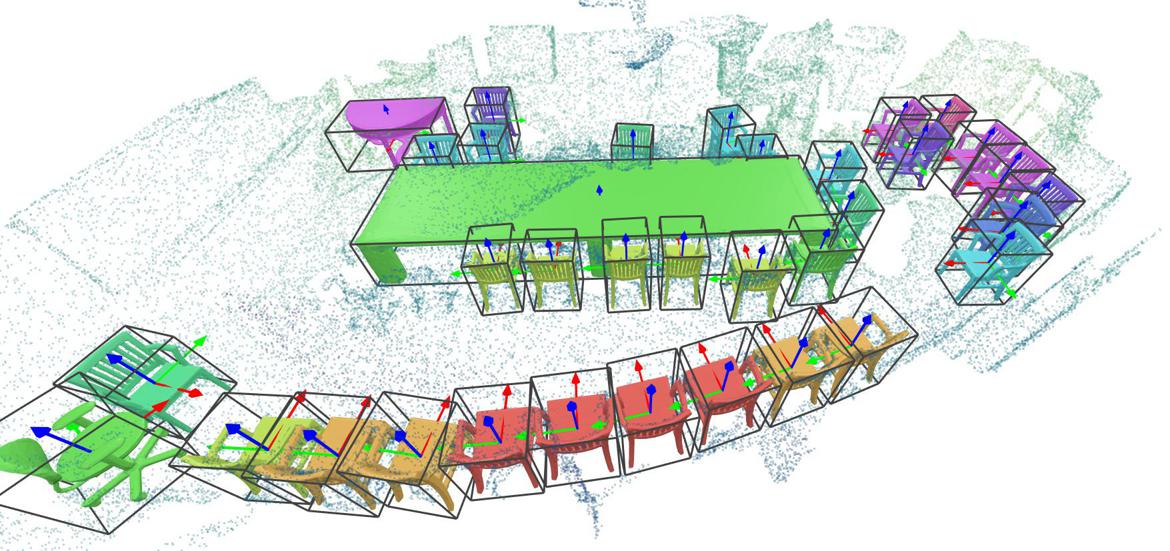}
		\includegraphics[width=\textwidth]
		{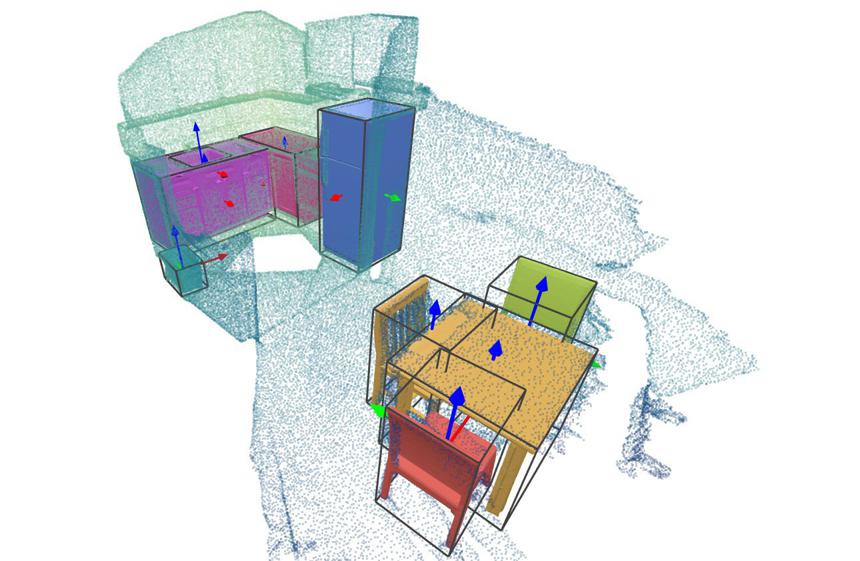}
		\includegraphics[width=\textwidth]
		{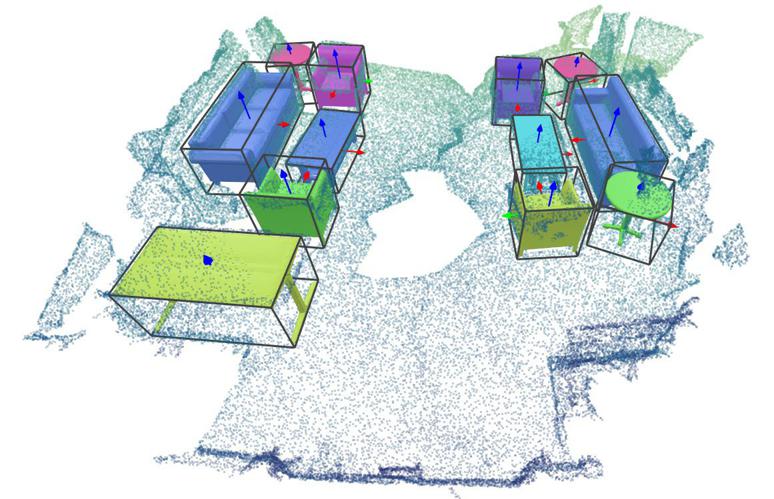}
		\caption{GT}
	\end{subfigure}
	\caption{Qualitative results of semantic instance reconstruction on ScanNet v2~\cite{dai2017scannet}. Note that RevealNet \cite{hou2020revealnet} preprocesses the scanned scenes into TSDF grids, while our method only uses the raw point clouds.}
	\label{fig:scene_recon}
	\vspace{-10pt}
\end{figure*}

\subsection{Comparisons to the State-of-the-Arts}
\label{sec:comp_to_sota}
\noindent\textbf{Qualitative Comparisons}. \label{comp:sota} We compare our method on semantic instance completion with the first and currently the state-of-the-art work, RevealNet \cite{hou2020revealnet}. Figure~\ref{fig:scene_recon} illustrates some qualitative results with different scene complexities on the test set of ScanNet (see more comparisons in the supplementary file). RevealNet discretizes the input scans into a volumetric grid that is preprocessed as a truncated signed distance field (TSDF). Their network is backboned with 3D CNN blocks to encode and decode the voxel occupancy of each object with a fully convolutional fashion. It can also back-project the 2D image features to TSDF grids to extend the representation of the input scans. In our method, only the geometry input is needed. The results in Figure~\ref{fig:scene_recon} show that our method presents better quality of object boxes and shape details. The reasons could be three-fold: 1. different with RevealNet which processes the whole scene with 3D CNNs, our method leverages the sparsity of point cloud and focuses on the object proposals with higher objectness. It enables us to ignore those empty spaces and saves computation load to reconstruct the objects of interest with high resolution (128-d); 2. voxelizing scenes into grids shows inadequacy in representing accurate object boxes (i.e., discrete box coordinates with axis-aligned orientations); 3. our shape generator presents joint effect that further improves the 3D detector and achieves better object localization.

We further quantitatively evaluate our method on single tasks as follows. Two configurations are considered to explore the complementary effects between the 3D detection and shape generation modules, i.e., training our network end-to-end (w/ joint) or training the 3D detector and shape generator individually with the other one fixed (w/o joint).

\begin{table*}[!h]
	\begin{center}
		\begin{tabular}{|l|c c c c c c c c |c|}
			\hline
			& display & bathtub & trashbin & sofa & chair & table & cabinet & bookshelf & mAP \\
			\hline
			\hline
			Inst Seg \cite{hou20193d} + Shape Comp \cite{dai2017shape}  & 2.27 & 1.14 & 1.68 & 14.86 & 9.93 & 3.90 & 7.11 & 3.03  & 5.49 \\
			Scan Comp \cite{dai2018scancomplete} + Inst Seg \cite{hou20193d} & 1.65 & 4.55 & 11.25 & 9.09 & 9.09 & 0.64 & 0.18 & 5.45  & 5.24 \\
			RevealNet \cite{hou2020revealnet} & 13.16 & 13.64 & 18.19 & \textbf{24.79} & \textbf{15.87} & \textbf{11.28} & 8.60 & 10.60 & 14.52 \\
			\hline
			\hline
			Ours (w/o joint) & 23.96 & 19.00 & 18.49 & 13.67 & 10.84 & 2.81 & 10.14 & 12.65 & 13.94\\
			Ours (w/ joint) & \textbf{26.67} & \textbf{27.57} & \textbf{23.34} & 15.71 & 12.23 & 1.92 & \textbf{14.48} & \textbf{13.39} & \textbf{16.90}\\
			\hline
		\end{tabular}
	\end{center}
	\vspace{-10pt}
	\caption{Comparisons on semantic instance completion. The results of (Inst Seg \cite{hou20193d} + Shape Comp \cite{dai2017shape}) and (Scan Comp \cite{dai2018scancomplete} + Inst Seg \cite{hou20193d}) are provided by \cite{hou2020revealnet}. The mAP scores are measured with the mesh IoU threshold at 0.5.}
	\label{compare:scenecomp}
	\vspace{-5pt}
\end{table*}

\begin{table}
	\begin{center}
		\begin{tabular}{|l|c|c|}
			\hline
			& Input & mAP \\
			\hline
			\hline
			3D-SIS \cite{hou20193d} & Geo+Image & 25.70 \\
			MLCVNet \cite{xie2020mlcvnet} & Geo Only & 33.40\\
			RevealNet \cite{hou2020revealnet} & Geo Only & 29.29\\
			\hline
			\hline
			Ours (w/o joint) & Geo Only & 32.63\\
			Ours (w/ joint) &  Geo Only & \textbf{35.10}\\
			\hline
		\end{tabular}
	\end{center}
	\vspace{-10pt}
	\caption{3D object detection on ScanNet v2. 3D-SIS \cite{hou20193d} and RevealNet \cite{hou2020revealnet} results are provided by the authors. MLCVNet results are retrained with the original network \cite{xie2020mlcvnet}. See per category scores in the supplementary file.}
	\label{compare:3ddetection}
	\vspace{-5pt}
\end{table}

\noindent\textbf{3D Object Detection.}
We evaluate our 3D detection by comparing with the prior art works, 3D-SIS \cite{hou20193d}, MLCVNet \cite{xie2020mlcvnet} and RevealNet \cite{hou2020revealnet} (see Table~\ref{compare:3ddetection}). 3D-SIS fuses the multi-view RGB images into the TSDF grids (by back-projecting 2D CNN features to 3D voxels) as inputs. It adapts the detection modality of Faster RCNN \cite{ren2015faster} from 2D image plane to 3D voxel grids. MLCVNet \cite{xie2020mlcvnet} extends VoteNet \cite{qi2019deep} by considering contextual information between objects. From the results, we observe that, with joint training, our shape generation module improves the 3D proposal network backbone (i.e., w/o joint training) on all categories (see per category scores in the supplementary file), and helps our method outperforms the state-of-the-arts, which is also shown consistently effective on other backbones (see analysis experiments).

\noindent\textbf{Object Reconstruction.}
We further evaluate the single object reconstruction quality with 3D mesh IoU, where the object shapes are predicted and evaluated in a canonical system (disentangled with 3D detection). Note that all the predictions and groundtruths in RevealNet \cite{hou2020revealnet} share the same coordinate system, i.e. the TSDF grid. RevealNet voxelizes 3D scenes with a uniform voxel size of $\approx4.7$ cm, which results in around $96\times48\times96$ voxels for a scene at the scale of $4.5\times2.25\times4.5$ meters. However, objects are independently reconstructed with the same resolution in our method. For a fair comparison, we present our results on different resolutions (16-d, 32-d, 64-d) and benchmark all methods with only geometric input (see the comparisons in Table~\ref{compare:objectrecon}). The results indicate that joint training the 3D detector with shape generator can even improve the performance on single object reconstruction. It implies that better box proposals will produce better spatial alignment and meaningful descriptions about the object property (e.g., size and category), which informs the shape generator to better approximate the target shapes. The results also demonstrate our mesh quality outperforms RevealNet with a large margin (over 11 points). Note that the resolution of 64-d even exceeds their scene resolution in vertical axis (48-d~\cite{hou2020revealnet}).

\begin{table}[!h]
	\begin{center}
		\begin{tabular}{|l|c|c|}
			\hline
			& resolution & 3D IoU \\
			\hline
			\hline
			RevealNet \cite{hou2020revealnet} & avg. 27-d & 20.48\\
			Ours (w/ \& w/o joint) & 16-d & \textbf{37.02} \& 35.75\\
			Ours (w/ \& w/o joint) & 32-d & \textbf{31.81} \& 30.21\\
			Ours (w/ \& w/o joint) & 64-d & \textbf{26.65} \& 24.97\\
			\hline
		\end{tabular}
	\end{center}
	\vspace{-10pt}
	\caption{Comparisons on object reconstruction. Per category scores are detailed in the supplementary file. }
	\label{compare:objectrecon}
	\vspace{-5pt}
\end{table}

\noindent\textbf{Semantic Instance Completion.}
We evaluate our method on scene completion at semantic-instance level. It measures how much the predicted object meshes overlay the ground-truths in a 3D scene. We compare our method with the state-of-the-art work, RevealNet \cite{hou2020revealnet}. We also follow their experiments by sequentially combing prior arts of instance segmentation \cite{hou20193d} with shape completion \cite{dai2017shape}, or combining scan completion (\cite{dai2018scancomplete}) with instance segmentation \cite{hou20193d}, for instance-level scene completion. As mentioned above, object meshes in our method are predicted with a higher resolution (128-d). For a fair comparison, we voxelize our scenes to the same resolution (4.7 cm per voxel) with them. The results are listed in Table~\ref{compare:scenecomp}. It shows that our joint method largely outperforms the decoupled methods and exceeds the state-of-the-art \cite{hou2020revealnet}. The comparison between w/ \& w/o joint training further explains the complementary effects between shape generation and 3D detection, which is inline with the former results that improving on each module will benefit the other one. For objects with very thin structure (e.g. tables), our method does not show advantage because a slight misalignment in 3D detection is more likely to affect their mesh overlay to the ground-truth in 3D scenes, although the object meshes are well reconstructed.

\begin{figure*}[!ht]
	\centering
	\begin{subfigure}[t]{0.24\textwidth}
		\includegraphics[width=\textwidth]
		{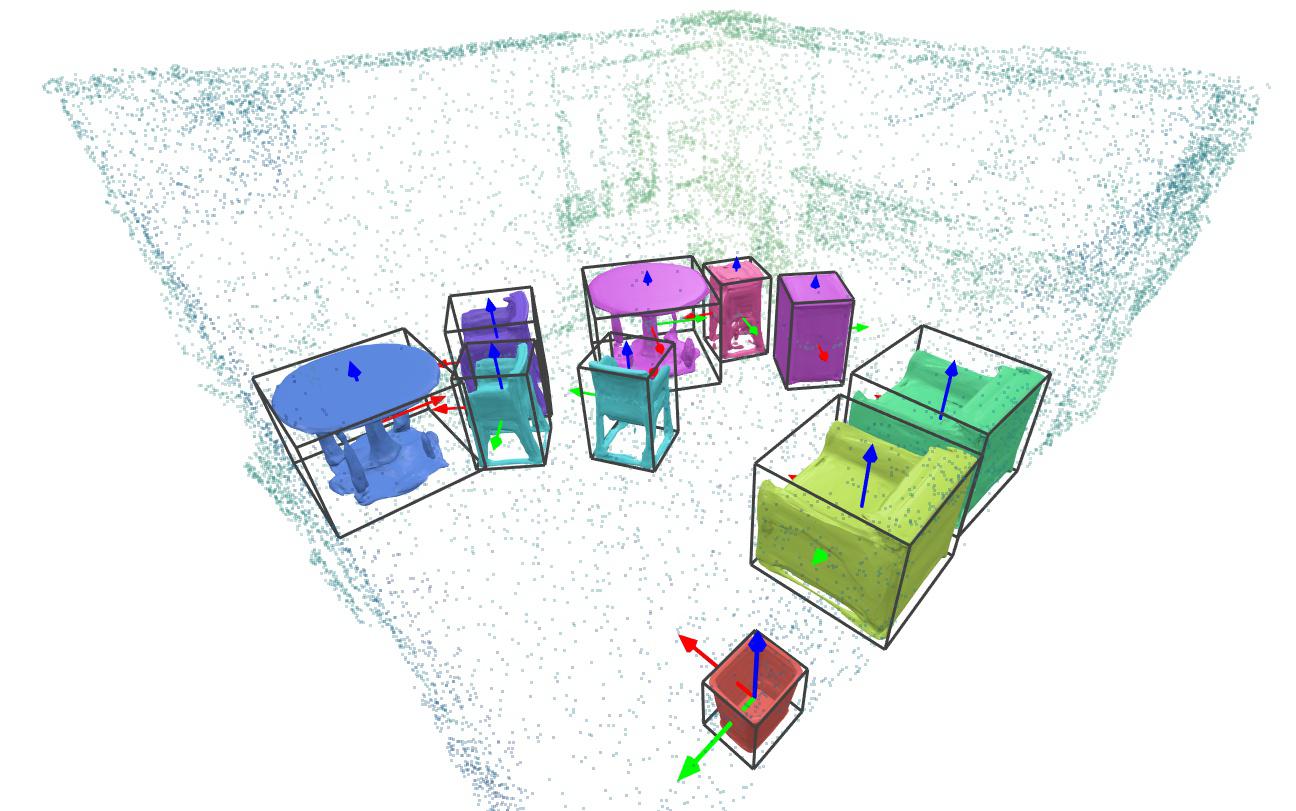}
		\includegraphics[width=\textwidth]
		{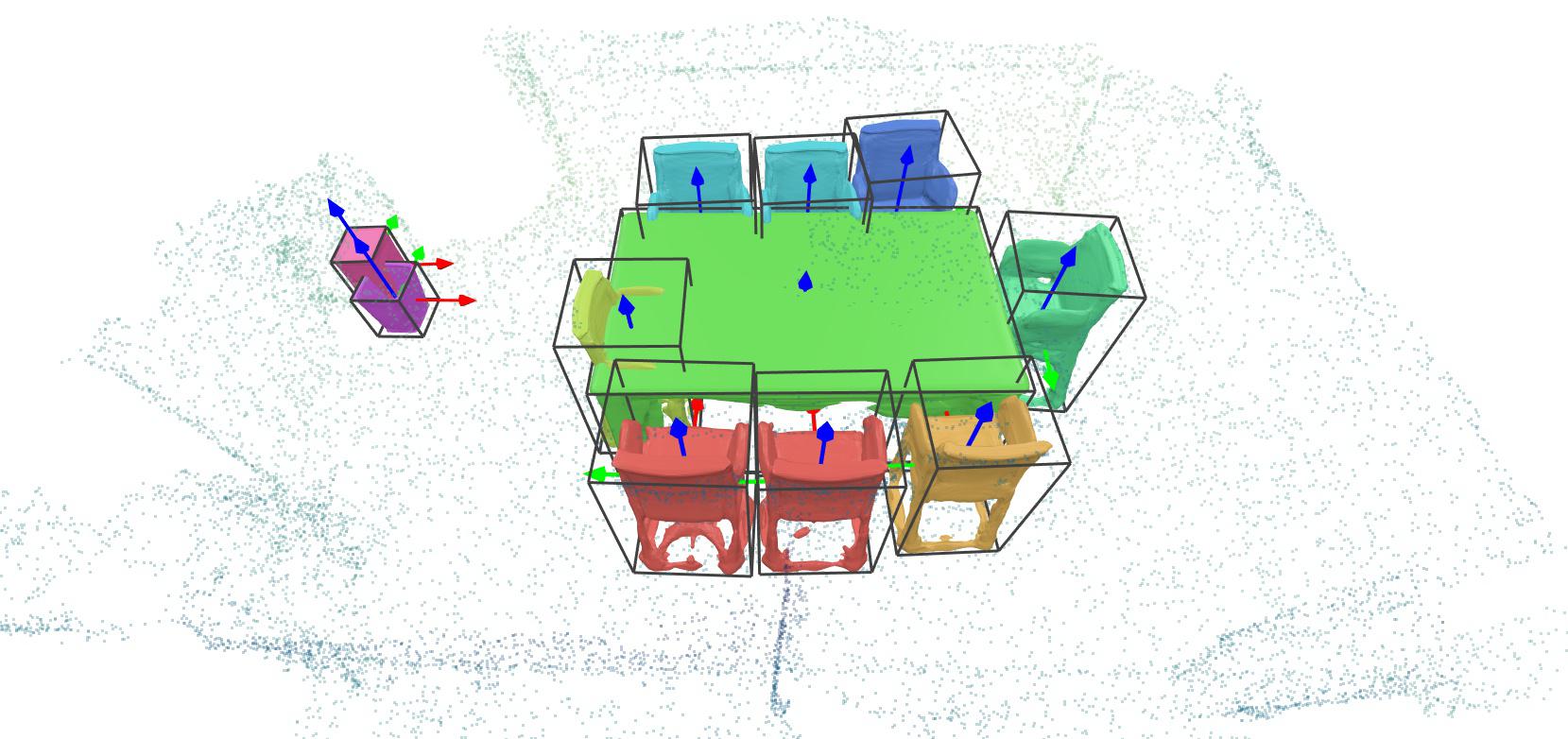}
		\caption{20K input points}
	\end{subfigure}
	\begin{subfigure}[t]{0.24\textwidth}
		\includegraphics[width=\textwidth]  
		{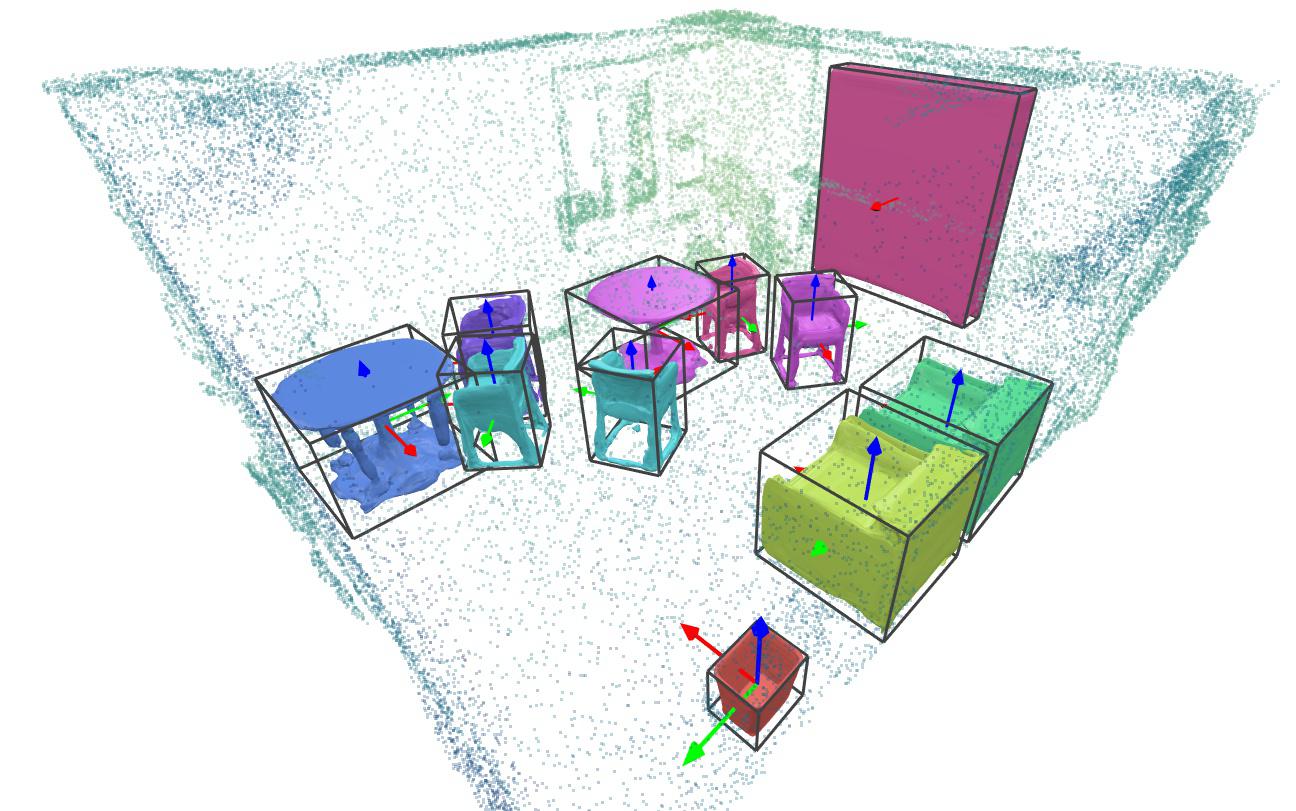}
		\includegraphics[width=\textwidth]  
		{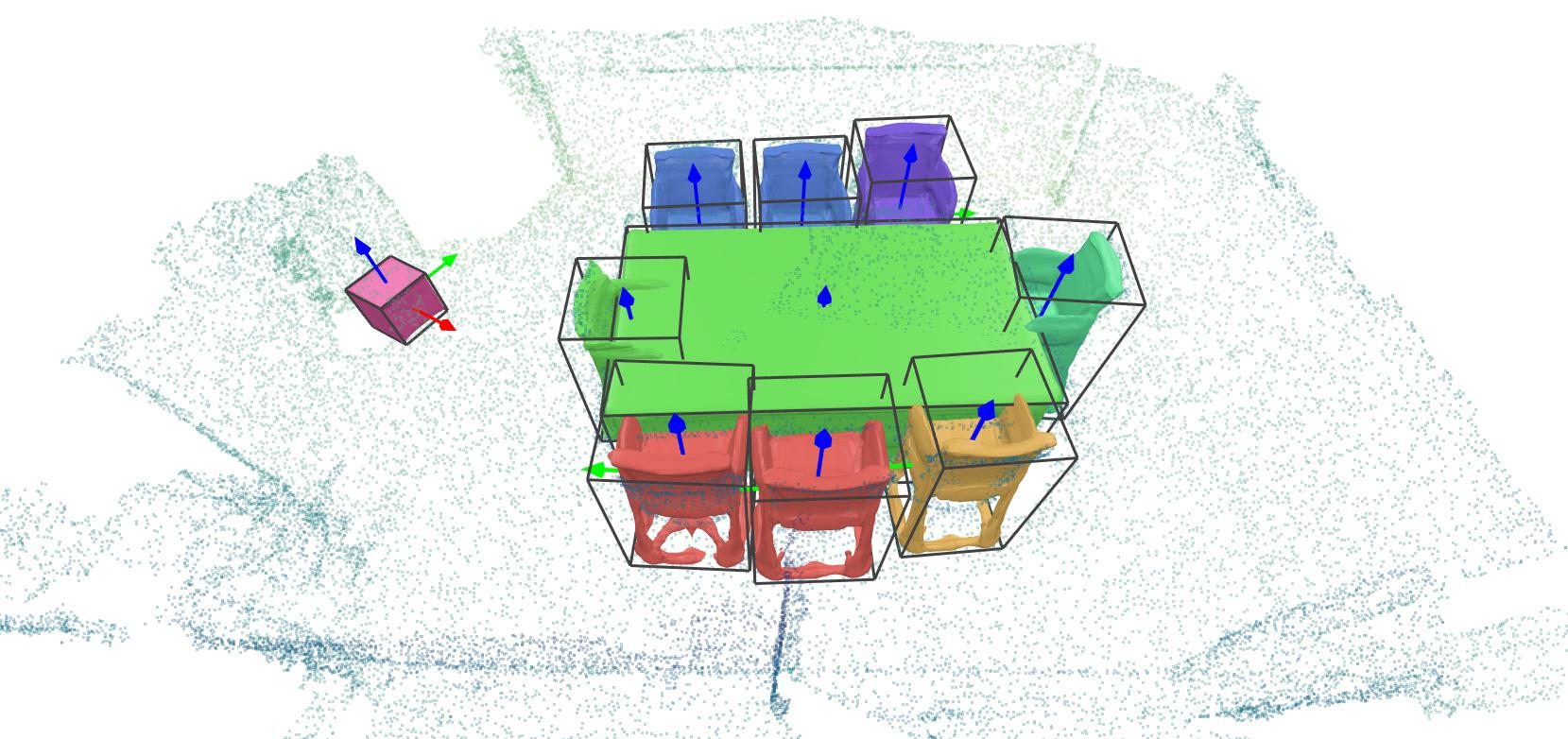}
		\caption{40K input points}
	\end{subfigure}
	\begin{subfigure}[t]{0.24\textwidth}
		\includegraphics[width=\textwidth]  
		{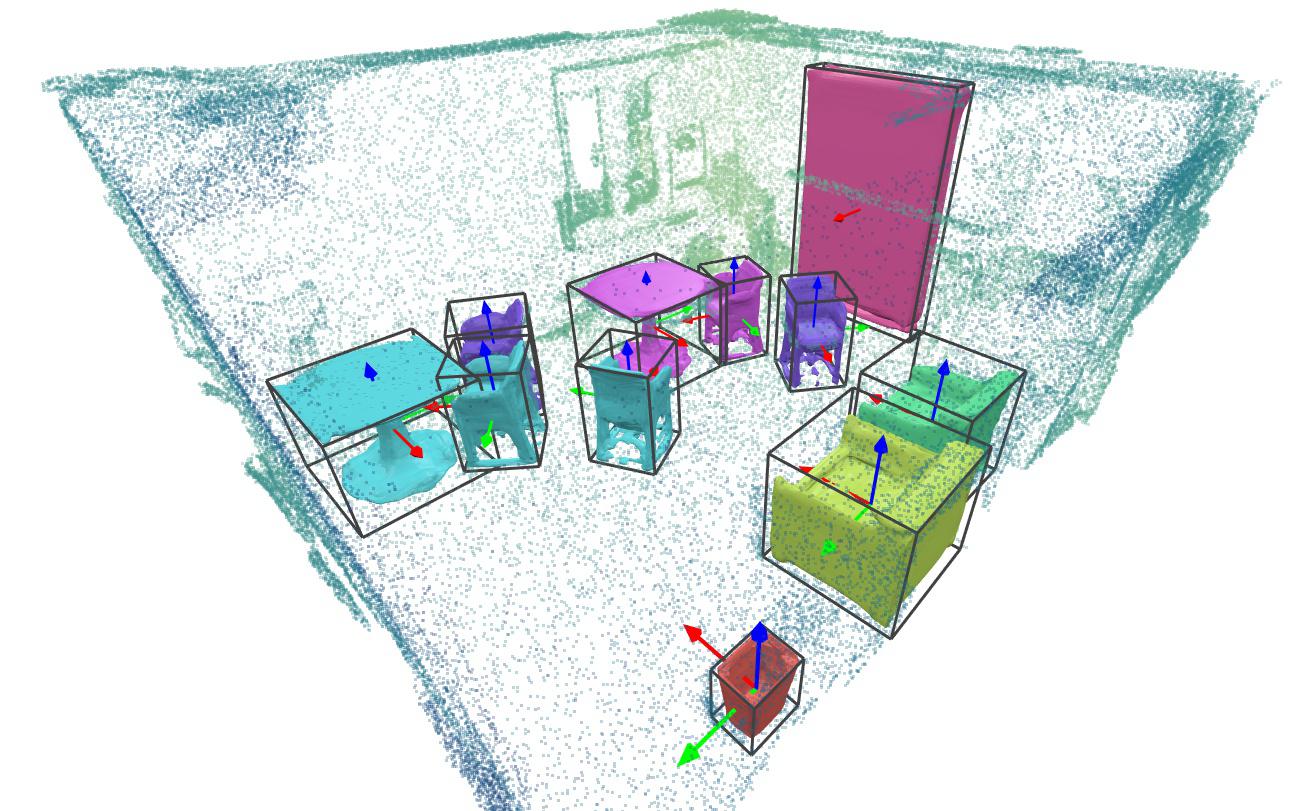}
		\includegraphics[width=\textwidth]  
		{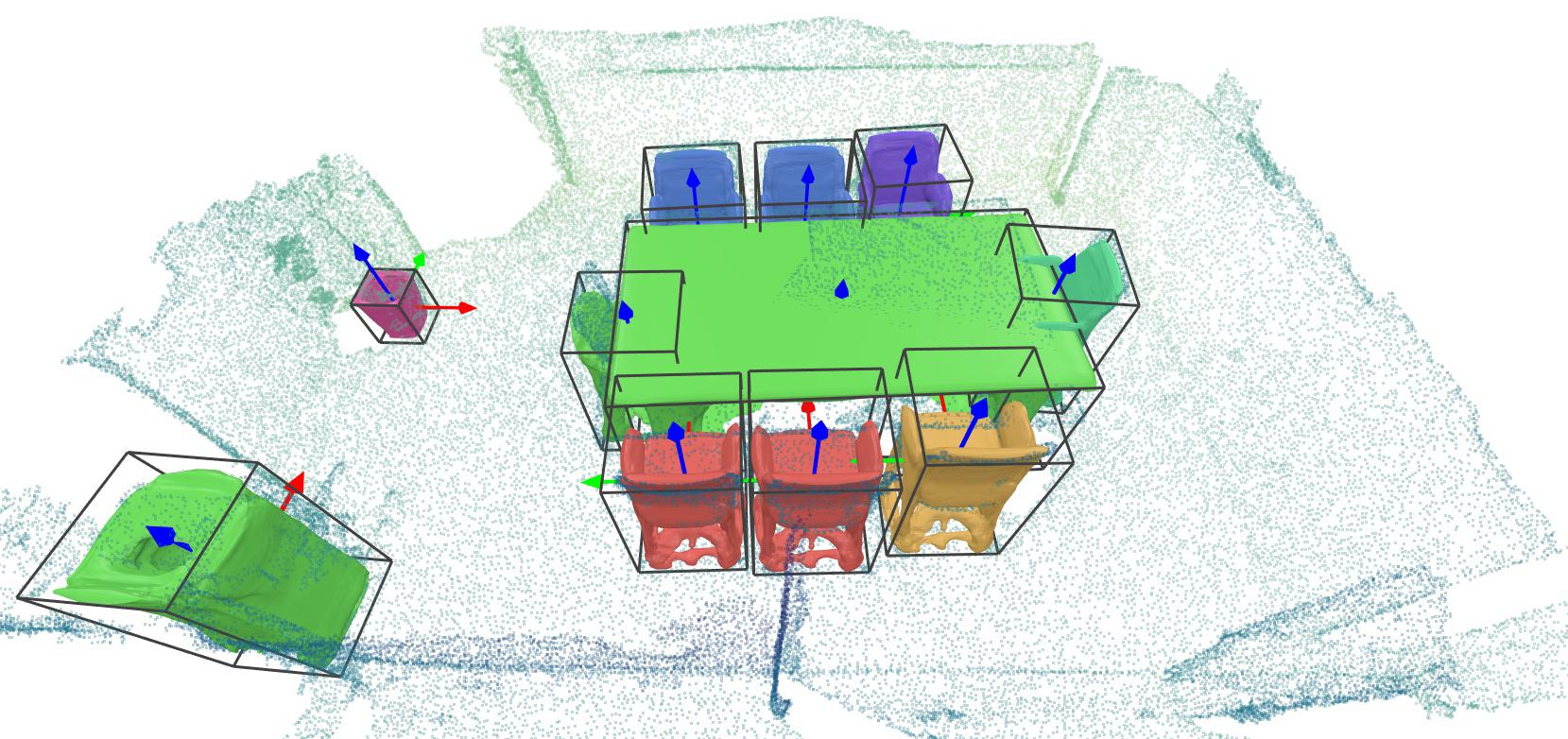}
		\caption{80K input points}
	\end{subfigure}
	\begin{subfigure}[t]{0.24\textwidth}
		\includegraphics[width=\textwidth]  
		{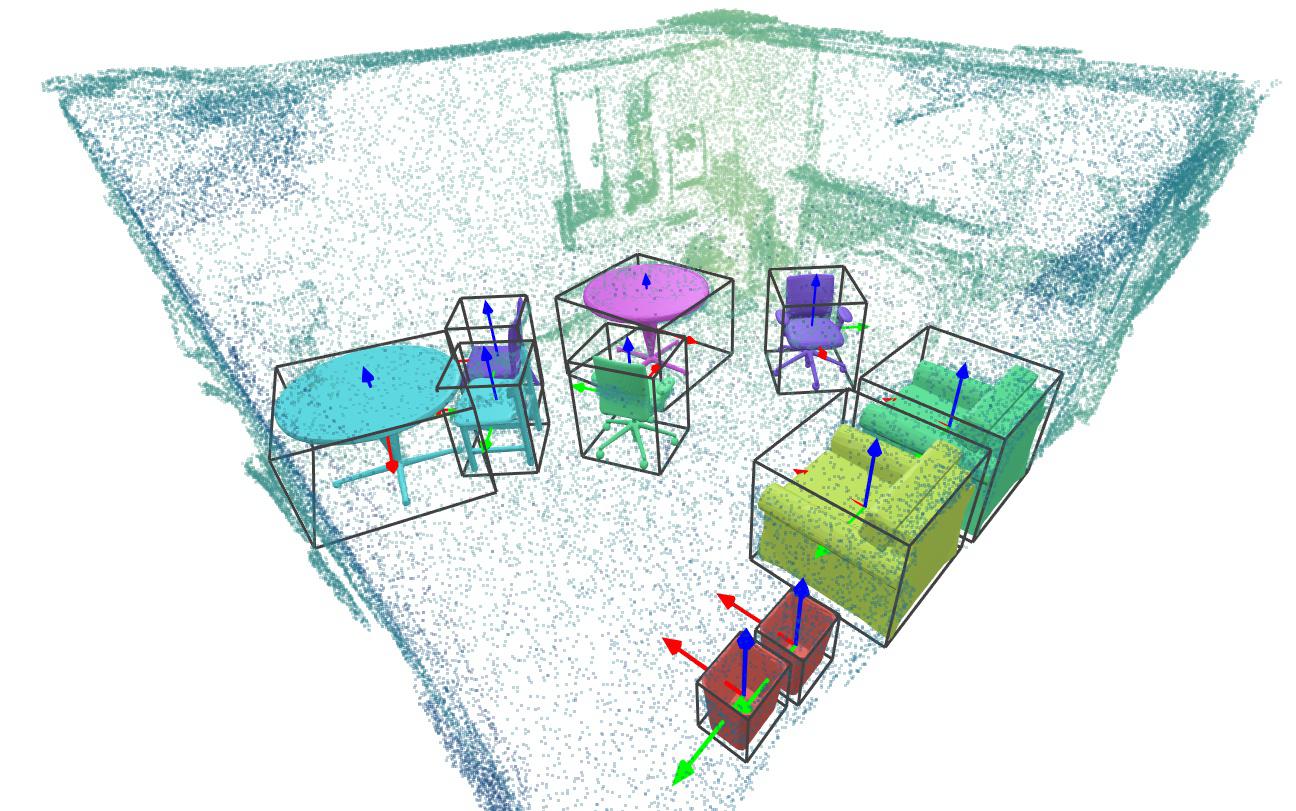}
		\includegraphics[width=\textwidth]  
		{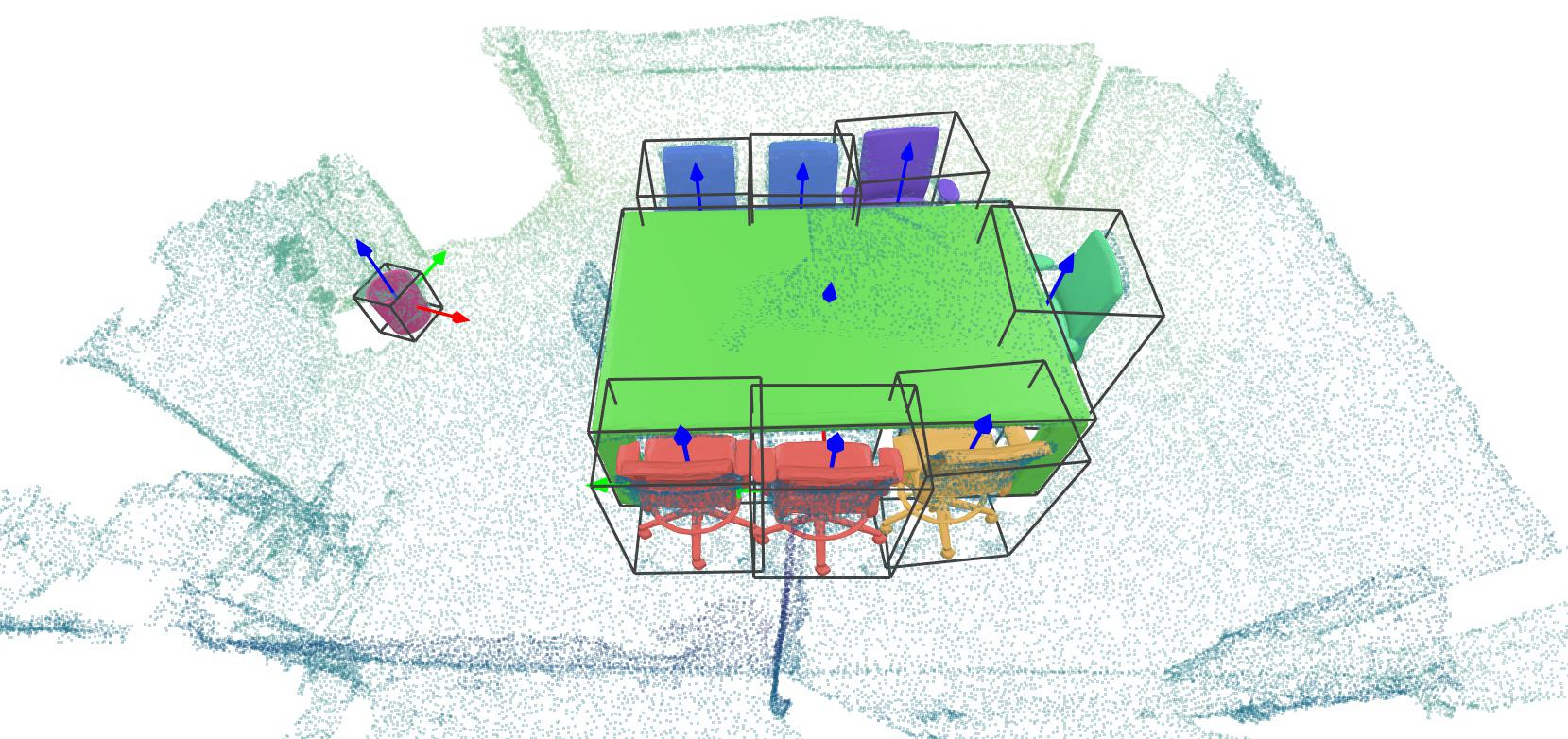}
		\caption{GT}
	\end{subfigure}
	\caption{Scene reconstruction with different point sparsity.}
	\label{fig:sparsity}
	\vspace{-10pt}
\end{figure*}

\subsection{Analysis Experiments}
\label{sec:analysis_experiments}
\noindent\textbf{Different Backbones.} In our network, we deploy the backbone of our 3D proposal network using VoteNet~\cite{qi2019deep}. We change our backbone to explore if the joint training still benefits other 3D detection networks. We consider in our experiments with the BoxNet \cite{qi2019deep}, MLCVNet \cite{xie2020mlcvnet} and VoteNet+Dynamic Graph CNN (DGCNN) \cite{dgcnn}. BoxNet directly predicts 3D boxes from the grouped points in point clouds. MLCVNet is a variant version of VoteNet that considers contextual information between objects. To this end, we also configure dynamic graph connections \cite{dgcnn} between votes to learn the relational features for object detection. The results in Table~\ref{compare:backbones} manifest that combing shape prediction in joint training consistently improves the 3D detection performance with different backbones.

\begin{table}[!h]
	\begin{center}
		\begin{tabular}{|l|c|c|}
			\hline
			& w/o joint & w/ joint \\
			\hline
			\hline
			BoxNet \cite{qi2019deep}  & 20.31 & 22.79 \\
			MLCVNet \cite{xie2020mlcvnet}  & 33.40 & 34.77 \\
			VoteNet+DGCNN \cite{dgcnn} & 29.88 & 32.30 \\
			\hline
			\hline
			Ours & 32.63 & 35.10\\
			\hline
		\end{tabular}
	\end{center}
	\vspace{-10pt}
	\caption{3D detection with different backbones.}
	\label{compare:backbones}
	\vspace{-5pt}
\end{table}

\noindent\textbf{Different Point Sparsity.}
Learning from point clouds makes our network adaptive to different scales of input points. We test the robustness of our method on different point sparsity, where our network is trained with 20K, 40K and 80K input points. The results in Figure~\ref{fig:sparsity} and Table~\ref{compare:sparity} show that our method still scan produce reasonable scenes with very sparse inputs (20K).

\begin{table}[!h]
	\begin{center}
		\begin{tabular}{|l|c|c|}
			\hline
			& 3D Detection & Instance Completion \\
			\hline
			\hline
			20K  & 33.89 & 14.39 \\
			40K  & 34.75 & 15.23 \\
			80K & 35.10 & 16.90 \\
			\hline
		\end{tabular}
	\end{center}
	\vspace{-10pt}
	\caption{Comparisons with different scales of input points.}
	\label{compare:sparity}
	\vspace{-5pt}
\end{table}

\noindent\textbf{Different Objectness Thresholds.} 
In section~\ref{sec:transformer}, we adopt \textit{Objectness Dropout} to keep proposal boxes with higher objectness (Top-$N_d$). It ensures meaningful object points as the input to learn shapes from those positive proposals. Thus, a high $N_d$ will involve in negative proposals (e.g. far away from any object) and undermines the shape learning, while a low $N_d$ will reduce the learning efficiency. We list the effects of using different $N_d$ in Table~\ref{compare:objectness}.

\begin{table}[!h]
	\begin{center}
		\begin{tabular}{|l|c|c|}
			\hline
			$N_d$ & 3D Detection & Instance Completion \\
			\hline
			\hline
			5  & 34.33 & 14.48 \\
			10  & \textbf{35.10} & \textbf{16.90} \\
			30 & 33.37 & 14.34 \\
			\hline
		\end{tabular}
	\end{center}
	\vspace{-10pt}
	\caption{Results on different thresholds for objectness dropout. $N_d$=10 is used in our network.}
	\label{compare:objectness}
	\vspace{-5pt}
\end{table}

\noindent\textbf{Different Skip Propagation.}
In section~\ref{sec:shapegeneration}, we propagate the proposal features $\left\{\bm{f}_{p}\right\}$ to point clusters $\{\tilde{\mathbf{P}}^{c}_{i}\}$ for shape generation. To investigate its effects, we configure extra three approaches of feature propagation for shape generation: 1. to decode shapes from proposal features without point clusters (\textit{c}\textsubscript{1}); 2. to learn shapes from the point clusters without proposal features (\textit{c}\textsubscript{2}); 3. to propagate the proposal features to point clusters but without the point denoizer (\textit{c}\textsubscript{3}). All these configurations are jointly trained. We list the comparisons in Table~\ref{compare:ablative}. The results indicate that the feature extraction for shape generation mainly affects the downstream shape generator thus further influence the instance completion, where combing the proposal features and point clusters with our skip propagation outcomes the best configuration.

\begin{table}[!h]
	\begin{center}
		\begin{tabular}{|l|c|c|}
			\hline
			& 3D Detection & Instance Completion \\
			\hline
			\hline
			\textit{c}\textsubscript{1} & 34.83 & 13.97 \\
			\textit{c}\textsubscript{2}  & 34.85 & 14.33 \\
			\textit{c}\textsubscript{3}  & 35.03 & 15.42 \\
			Full & \textbf{35.10} & \textbf{16.90} \\
			\hline
		\end{tabular}
	\end{center}
	\vspace{-10pt}
	\caption{Ablation on skip propagation.}
	\label{compare:ablative}
	\vspace{-5pt}
\end{table}

\section{Conclusion}
We propose a novel learning modality, namely \textit{RfD-Net}, for semantic instance reconstruction directly point clouds. It disentangles the problem with a \textit{reconstruction-from-detection} manner. Instance shapes are predicted with global object localization and local shape prediction, which are connected with a spatial transformer and a skip propagation module to bridge the information flow from shapes to detections. It facilitates the complementary effects and jointly improves the performance of 3D detection and shape generation. The experiments further demonstrate that our method achieves much better mesh quality in scene reconstruction and outperforms the state-of-the-art in object reconstruction, 3D detection and semantic instance completion.

\section*{Acknowledgements}

\begin{footnotesize}
	
This project was supported in part by the National Key R\&D Program of China with grant No.~2018YFB1800800, the Key Area R\&D Program of Guangdong Province with grant No.~2018B030338001, Guangdong Research Project No.~2017ZT07X152, the National Natural Science Foundation of China 61902334, Shenzhen Fundamental Research (General Project) JCYJ20190814112007258, a TUM-IAS Rudolf Mößbauer Fellowship, the ERC Starting Grant Scan2CAD (804724), the German Research Foundation (DFG) Grant Making Machine Learning on Static and Dynamic 3D Data Practical, Bournemouth University and China Scholarship Council. We also would like to thank Prof.~Angela Dai for the video voice over in our work.
	
\end{footnotesize}

{\small
\bibliographystyle{ieee_fullname}
\bibliography{paper}
}

\newpage
\begin{appendix}
\appendix
\section*{Appendix}
\section{Network and Layer Specifications}
In this section, we provide all the parameters, layer specifications and weights used in loss functions. We uniformly denote the fully-connected layers by $\text{MLP}\left[l_{1},...,l_{d}\right]$, where $l_{i}$ is the number of neurons in the $i$-th layer.

\subsection{3D Detector}
In section 3.1, we predict object proposals from $N$ input points with VoteNet \cite{qi2019deep} as the backbone. It produces $N_{p}$ proposals with $D_{p}$-dim features (i.e. proposal features $\bm{F}_{p}\in\mathbb{R}^{N_{p}\times D_{p}}$ in our paper), from which we regress the $D_{b}$-dim box parameters with $\text{MLP}\left[128,128,69\right]$ ($N$=80K, $N_{p}$=256, $D_{p}$=128, $D_{b}$=69). As in \cite{qi2019deep}, the 69-dim box parameters are encoded by center $\bm{c}\in\mathbb{R}^{3}$, scale $\bm{s}^{3}\in\mathbb{R}^{3}$, heading angle $\theta\in\mathbb{R}$, semantic label $l$, and objectness score $s_{obj}$. $s_{obj}$ is a probability value indicating whether the proposal is close to ($<$0.3 meter, positive) or
far from ($>$0.6 meter, negative) any ground-truth object center.

\subsection{Spatial Transformer}
\noindent\textbf{Objectness Dropout Layer.}
In section 3.2, we adopt an \textit{Objectness Dropout} layer to reserve the Top-$N_{d}$ proposals with higher objectness ($N_{d}=10$) for shape learning. In test, we replace it with 3D Non Maximum Suppression (3D NMS) to produce the output boxes and corresponding shapes with the 3D Box IoU threshold of 0.25 in evaluation.

\begin{figure}[!h]
	\centering
	\includegraphics[width=1\linewidth]{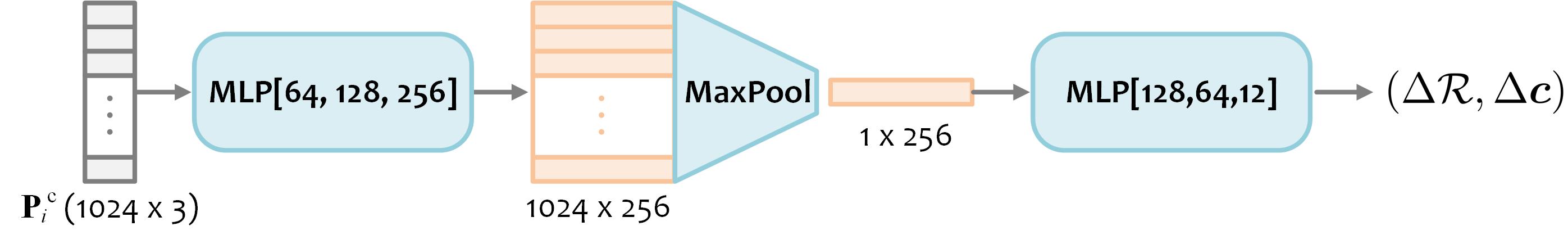}
	\caption{Rotation and translation adjustment.}
	\label{fig:groupalign}
	\vspace{-5pt}
\end{figure}

\noindent\textbf{Group \& Align.} From the $N_{d}$ box proposals, we group the neighboring $M_{p}$ points that are located within a radius $r$ to each box center using a group layer \cite{qi2017pointnet++}. $M_{p}$=1024, $r$=1. It produces $N_{d}$ point clusters $\left\{\mathbf{P}^{c}_{i}\right\}$ ($i=1,2,...,N_{d}, \mathbf{P}^{c}_{i}\in\mathbb{R}^{M_{p}\times3}$). After grouping, we align the 3D points in each cluster to a canonical system with the Equation~1 in our paper, where the rotation and translation adjustment $\left(\Delta\mathcal{R},\Delta\bm{c}\right)$ are predicted from $\mathbf{P}^{c}_{i}$ as in Figure~\ref{fig:groupalign}. $\left(\Delta\mathcal{R},\Delta\bm{c}\right)$ are predicted without supervision, which asks for the network to search for the optimal adjustment in spatial alignment (see Equation~1 in our paper).

\subsection{Shape Generator}
In section~3.3, we design the shape generator with two parts (see Figure~3 of the paper): a shape encoder extended with skip propagation; and a shape decoder based on conditional batch normalization.

\begin{figure}[!h]
	\centering
	\includegraphics[width=1\linewidth]{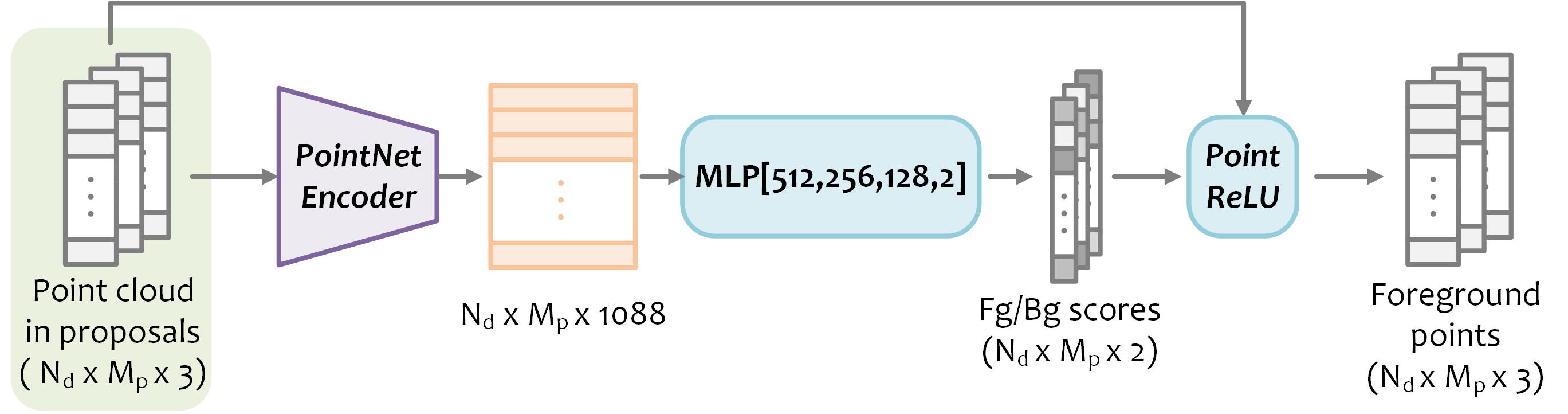}
	\caption{Denoiser in skip propagation.}
	\label{fig:denoiser}
	\vspace{-5pt}
\end{figure}

\begin{figure*}[!t]
	\centering
	\includegraphics[width=1\linewidth]{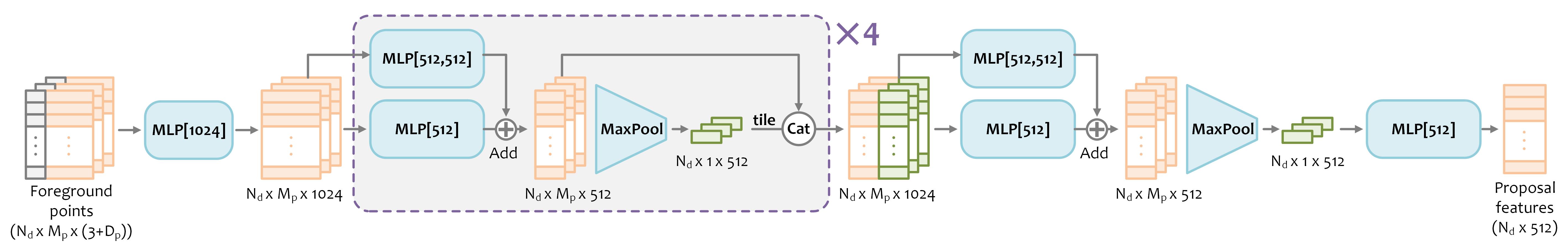}
	\caption{PointNet-based shape encoder with residual connection.}
	\label{fig:shapeencoder}
	\vspace{-5pt}
\end{figure*}

\noindent\textbf{Shape Encoder.}
The layer specification of the denoiser in skip propagation is illustrated in Figure~\ref{fig:denoiser}. The PointNet encoder \cite{qi2017pointnet} designed with residual connection is shown in Figure~\ref{fig:shapeencoder}. It takes the extended point clusters as input (see section~3.3 in the paper) and outputs the new proposal features $\left\{\bm{f}^{*}_{p}\in\mathbb{R}^{D_{s}}\right\}$, $D_{s}=512$ to decode shapes.

\begin{figure*}[!t]
	\centering
	\includegraphics[width=1\linewidth]{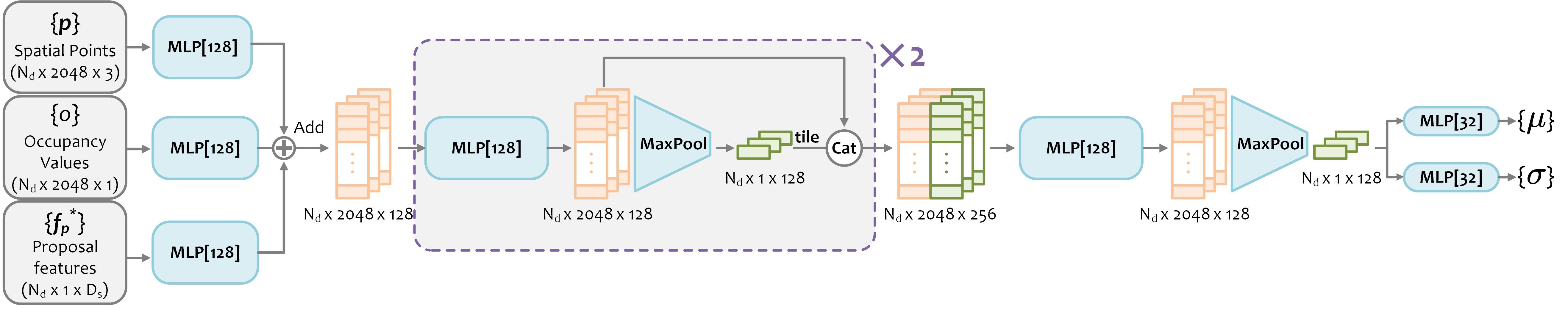}
	\caption{Latent encoder for probabilistic shape generation.}
	\label{fig:latentencoder}
	\vspace{-5pt}
\end{figure*}

\begin{figure*}[!t]
	\centering
	\includegraphics[width=1\linewidth]{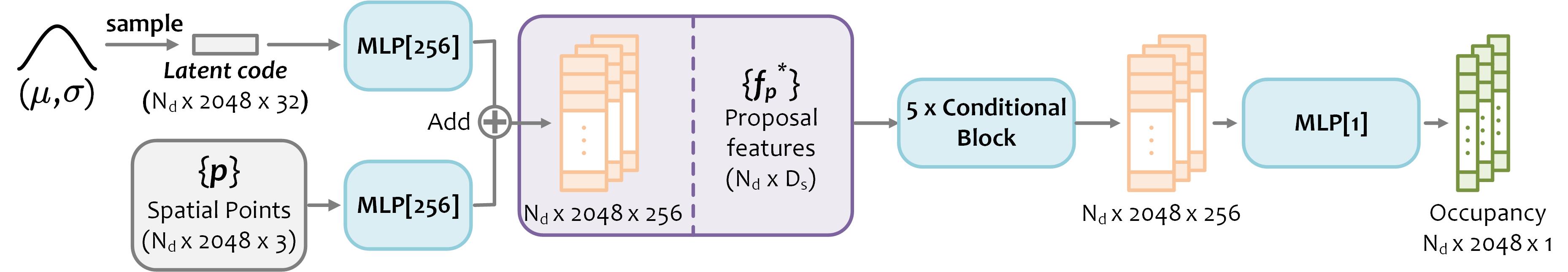}
	\caption{Shape decoder with Conditional Batch Normalization layers.}
	\label{fig:decoder}
	\vspace{-5pt}
\end{figure*}

\noindent\textbf{Shape Decoder.} We build the shape decoder as a probabilistic generative model. The latent encoder \cite{mescheder2019occupancy} is fed with the proposal feature $\left\{\bm{f}^{*}_{p}\right\}$, spatial points $\left\{\bm{p}\in\mathbb{R}^{3}\right\}$ with corresponding occupancy values $\left\{o\right\}$. It outputs the mean and standard deviation $\left(\bm{\mu}, \bm{\sigma}\right)$ to approximate the standard normal distribution. We illustrate the latent encoder in Figure~\ref{fig:latentencoder}, where the 2,048 spatial points are randomly sampled in the shape bounding cube. In our method, we separately sample 1,024 points inside and 1,024 points outside the shape mesh. From the predicted distribution $N\left(\bm{\mu}, \bm{\sigma}\right)$, we sample a latent code $\bm{z}\in\mathbb{R}^{L}$ ($L=32$) to predict the occupancy values $\left\{o\right\}$ from $\left\{\bm{p}\right\}$ conditioned on $\bm{f}^{*}_{p}$ for each object. The shape decoder is based on the Conditional Batch Normalization \cite{de2017modulating,dumoulin2016adversarially} layers, which is illustrated in Figure~\ref{fig:decoder}.

During test, we directly initialize the latent codes with zeros. As discussed in section~3.3 of our paper, we uniformly sample spatial points in the object bounding box, and use Marching Cubes \cite{lorensen1987marching} to extract the iso-surface as object meshes. Specifically, we adopt the efficient Multi-resolution Iso-Surface Extraction (MISE) algorithm \cite{mescheder2019occupancy} to improve the spatial sampling efficiency and extract meshes under 128-d occupancy grids.

\subsection{Weights in Loss Functions}
We list the weights to balance different loss functions as follows. We set $\lambda_{cls}=0.1$ as the hybrid ratio of combing the classification and regression losses, i.e., $ \lambda_{cls}\mathcal{L}_{cls} + \mathcal{L}_{reg}$, in defining the scale loss $\mathcal{L}_{s}$ and heading angle loss $\mathcal{L}_{\theta}$. Then the box loss $\mathcal{L}_{box}$ in our paper can be denoted as:
\begin{equation}
	\mathcal{L}_{box} = \mathcal{L}_{v} + \mathcal{L}_{c} + \mathcal{L}_{s} + \mathcal{L}_{\theta} + \lambda_{l}\mathcal{L}_{l} + \lambda_{obj}\mathcal{L}_{obj},
\end{equation}
where $\lambda_{obj}=0.5$, $\lambda_{l}=0.1$.
For shape loss in Equation~3 of our paper, we set $\lambda_{seg}=1e2$. Then the total loss for end-to-end training can be summarized as:
\begin{equation}
	\mathcal{L} = \mathcal{L}_{box} + \lambda\mathcal{L}_{shape},
\end{equation}
where $\lambda=5e$-3.

\section{Efficiency and Memory in Inference}
We train our network with two NVIDIA TITAN-Xp GPUs and test it on a single GPU. We also compare the inference timing in single forward pass and GPU memory usage with \cite{hou2020revealnet} (see Table~\ref{compare:efficiency}). It shows that our method have comparable efficiency with the state-of-the-art but requires much fewer memory ($\approx$1/3), which indicates that learning shapes directly from the raw point cloud consumes fewer computation resource and hardware requirement than processing 3D scenes with TSDF grids.

\begin{table}[!h]
	\begin{center}
		\begin{tabular}{|l|c|c|}
			\hline
			& Max. Time (s) & Max. Memory (MB)\\
			\hline
			\hline
			RevealNet \cite{hou2020revealnet} & 0.72 & 4273 \\
			Ours & 0.68 & 1239 \\
			\hline
		\end{tabular}
	\end{center}
	\caption{Maximal inference time (seconds) and GPU memory (MB) of a single forward pass in ScanNet v2~\cite{dai2017scannet}.}
	\label{compare:efficiency}
\end{table}

\begin{table*}
	\begin{center}
		\begin{tabular}{|l|c|c c c c c c c c|c|}
			\hline
			& Input & table & bathtub & trashbin & sofa & chair & cabinet & bookshelf & display & mAP \\
			\hline
			\hline
			3D-SIS \cite{hou20193d} & Geo+RGB & 42.76 & 3.03 & 20.04 & 34.39 & 63.26 & 21.54 & 16.92 & 3.69 &  25.70 \\
			MLCVNet \cite{xie2020mlcvnet} & Geo Only & 44.51 & 22.39 & 10.10 & \textbf{53.13} & 78.74 & 26.34 & 22.93 & 8.90 & 33.40\\
			RevealNet \cite{hou2020revealnet} & Geo Only & 35.64 & 14.94 & 26.77 & 29.96 & 53.18 & 26.63 & 15.89 & 31.30  & 29.29\\
			\hline
			\hline
			Ours (w/o joint) & Geo Only & 46.67 & 19.09 & 15.30 & 51.71 & 77.22 & 24.62 & 18.58 & 7.03  & 32.63\\
			Ours (w/ joint) &  Geo Only & \textbf{49.90} & \textbf{23.63} & 15.69 & 52.34 & \textbf{79.88} & 26.72 & \textbf{23.20} & 9.23 & \textbf{35.10}\\
			\hline
		\end{tabular}
	\end{center}
	\caption{3D object detection on ScanNet v2. 3D-SIS \cite{hou20193d} and RevealNet \cite{hou2020revealnet} results are provided by the authors. MLCVNet results are retrained with the original network \cite{xie2020mlcvnet}. Scores above are evaluated with mAP@0.5.}
	\label{compare:3ddetection_supp}
\end{table*}

\begin{table*}
	\begin{center}
		\begin{tabular}{|l|c|c c c c c c c c|c|}
			\hline
			& resolution & table & bathtub & trashbin & sofa & chair & cabinet & bookshelf & display & 3D IoU \\
			\hline
			\hline
			RevealNet \cite{hou2020revealnet} & avg. 27-d & 17.43 & 12.64 & 17.90 & 28.73 & 29.61 &  20.78 & 18.05 & 18.68 & 20.48 \\
			\hline
			\hline
			Ours (w/o joint) & 16-d & 16.08 & 32.74 & 36.24 & 46.51 & 35.53 & 45.71 & 33.55 & 39.63 & 35.75\\
			Ours (w/o joint) &  32-d & 11.65 & 28.07 & 35.86 & 39.97 & 28.89 & 44.27 & 23.54 & 29.44 & 30.21\\
			Ours (w/o joint) & 64-d & 20.68 & 29.46 & 25.43 & 24.19 & 23.55 & 30.84 & 22.43 & 23.20 & 24.97\\
			\hline
			\hline
			Ours (w/ joint) & 16-d & 19.22 & 32.55 & 36.92 & 46.73 & 37.05 & 49.25 & 35.14 & 39.28 & 37.02 \\
			Ours (w/ joint) &  32-d & 15.24 & 28.55 & 36.09 & 41.47 & 30.91 & 47.03 & 24.94 & 30.27 & 31.81 \\
			Ours (w/ joint) &  64-d & 22.12 & 29.24 & 28.12 & 27.80 & 26.05 & 33.75 & 23.27 & 22.87 & 26.65\\
			\hline
		\end{tabular}
	\end{center}
	\caption{Comparisons on object reconstruction. Scores above are measured with 3D mesh IoU.}
	\label{compare:object_recon_supp}
\end{table*}

\section{Detailed Quantitative Comparisons}
In this section, we list the per-category scores of 3D detection and object reconstruction in Table~\ref{compare:3ddetection_supp} and Table~\ref{compare:object_recon_supp} (refer to section~5.2 in the paper).

\section{More Qualitative Comparisons}
\begin{figure*}
	\centering
	\begin{subfigure}[t]{0.24\textwidth}
		\includegraphics[width=\textwidth]  
		{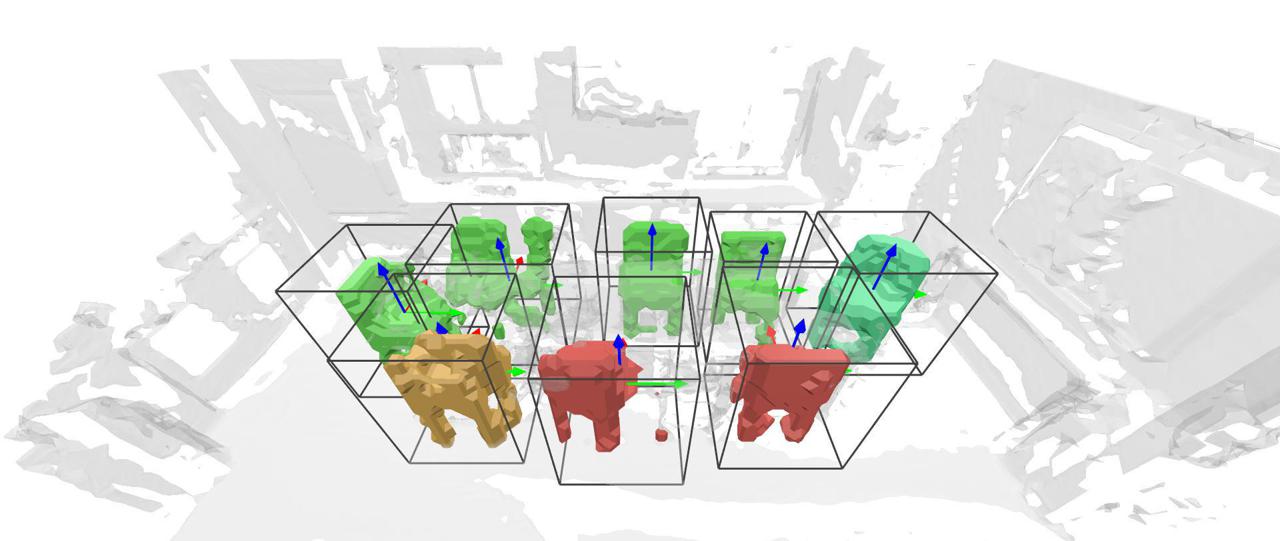}
		\includegraphics[width=\textwidth]  
		{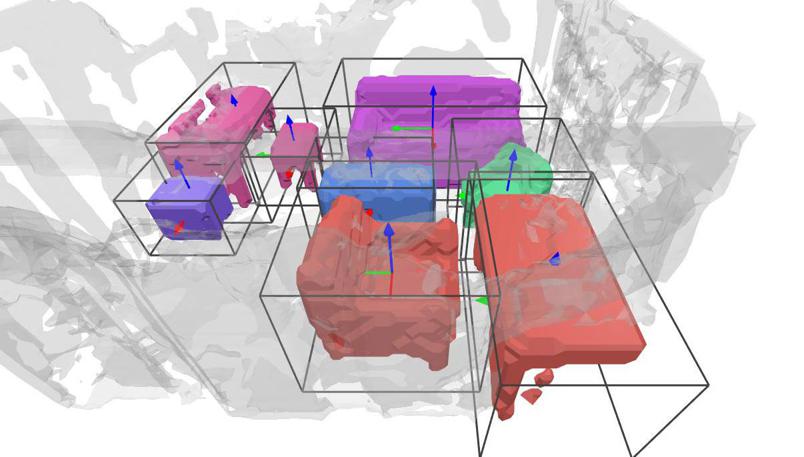}
		\includegraphics[width=\textwidth]  
		{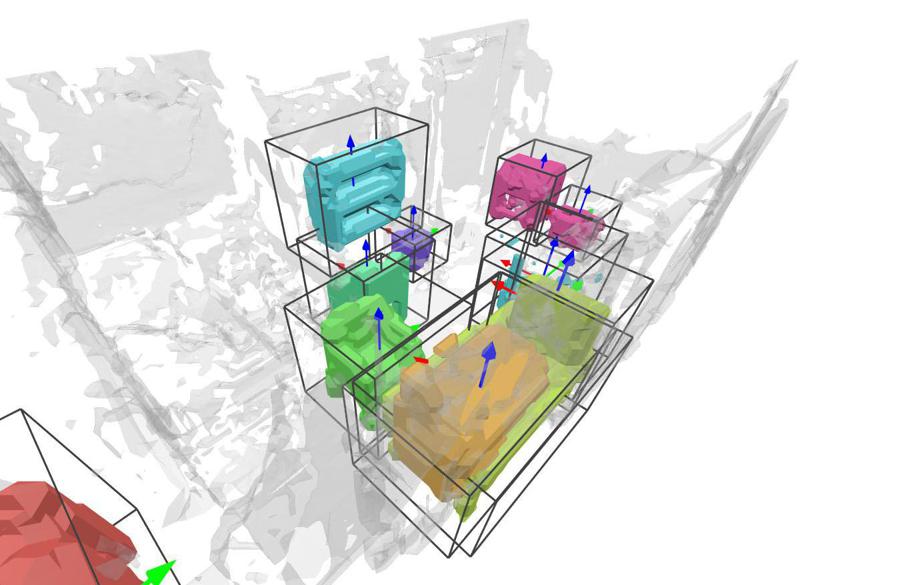}
		\includegraphics[width=\textwidth]  
		{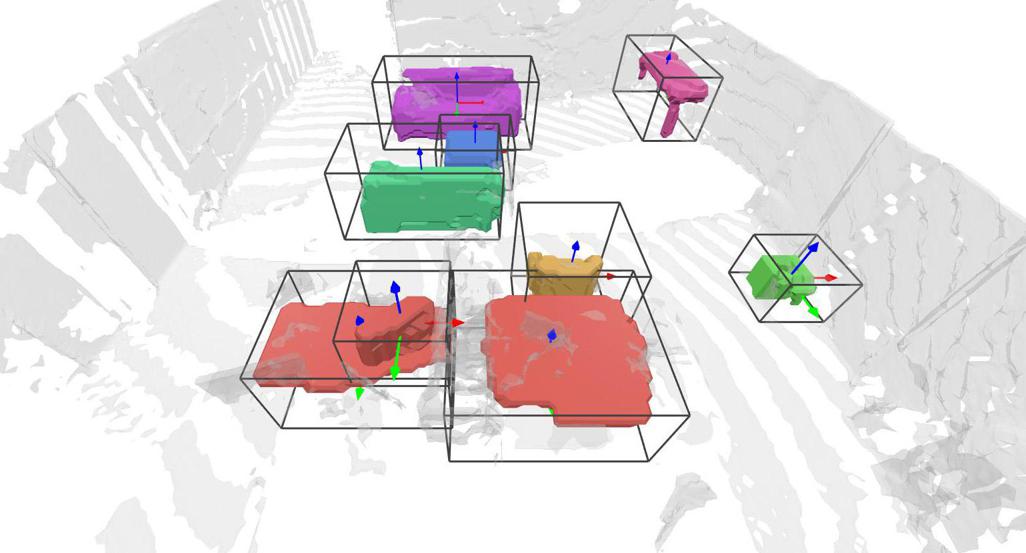}
		\includegraphics[width=\textwidth]  
		{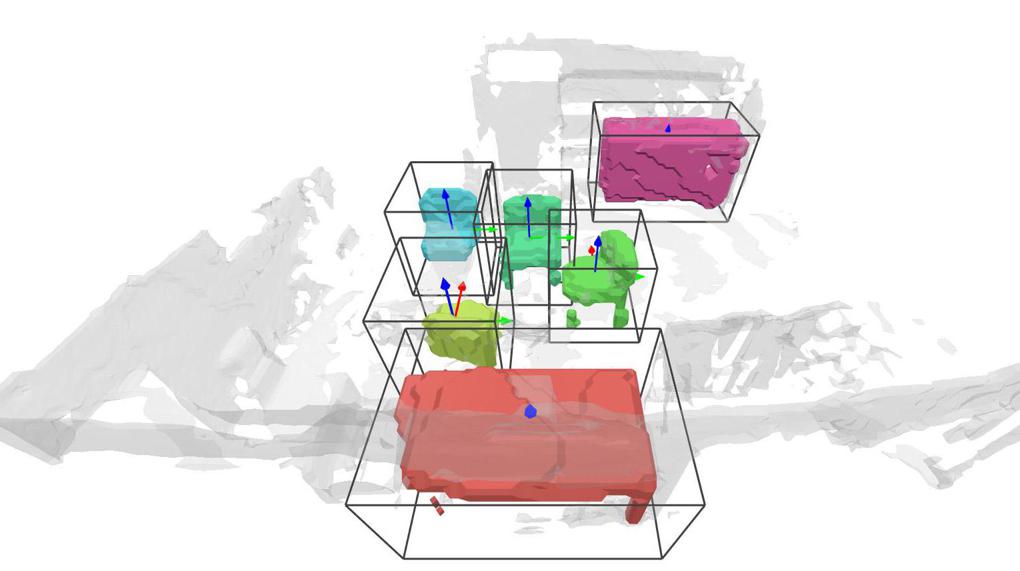}
		\includegraphics[width=\textwidth]  
		{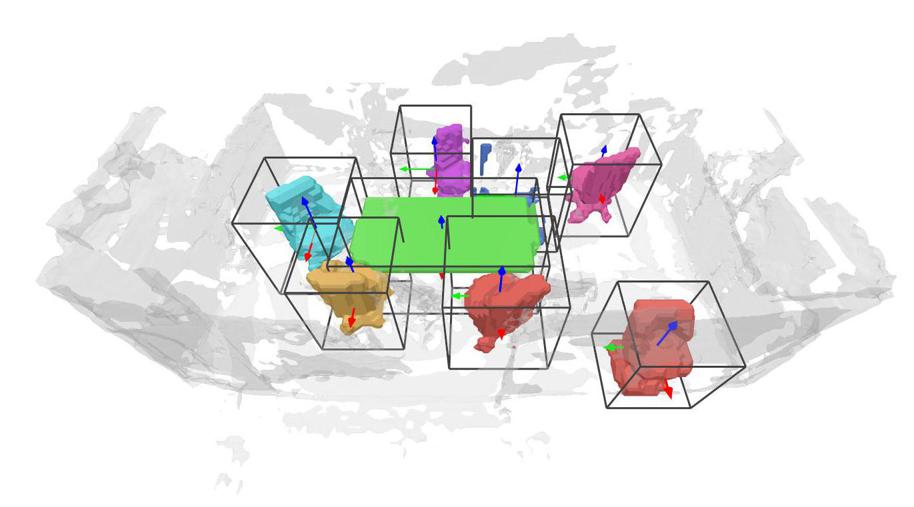}
		\includegraphics[width=\textwidth]  
		{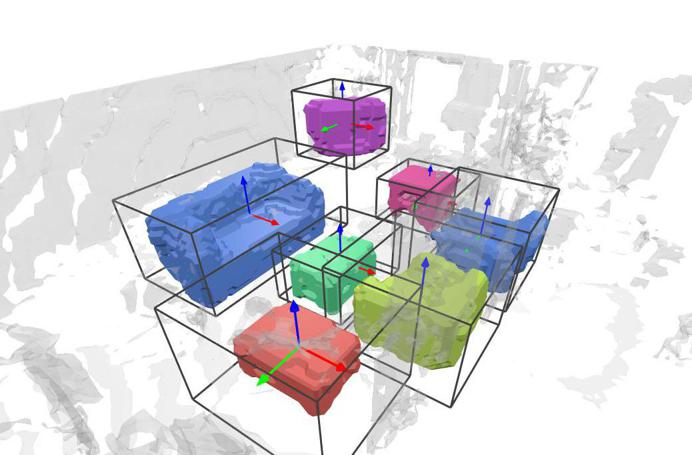}
		\includegraphics[width=\textwidth]  
		{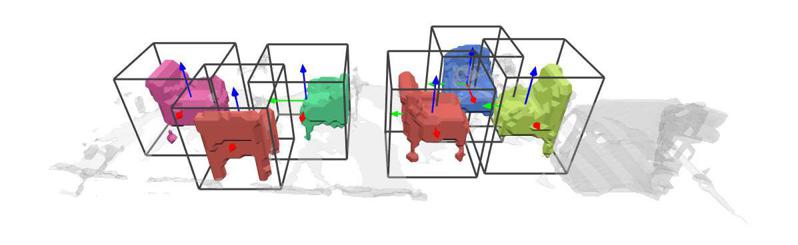}
		\includegraphics[width=\textwidth]  
		{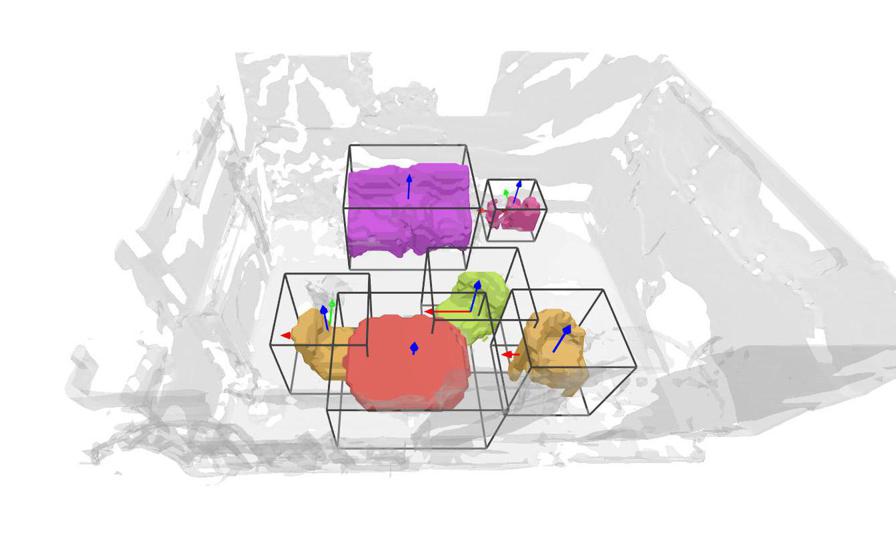}
		\includegraphics[width=\textwidth]  
		{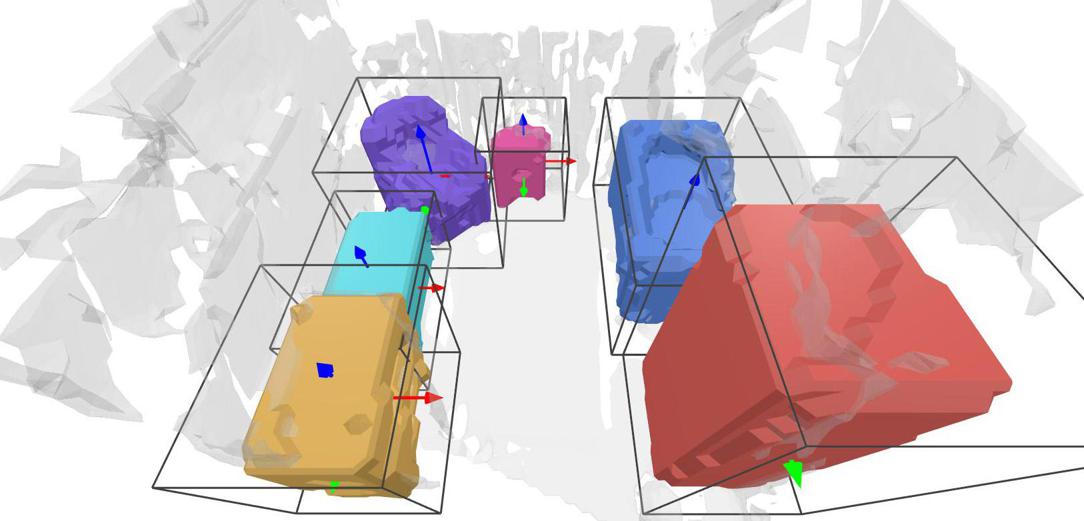}
		\caption{RevealNet \cite{hou2020revealnet} (Geo+Image)}
	\end{subfigure}
	\begin{subfigure}[t]{0.24\textwidth}
		\includegraphics[width=\textwidth]
		{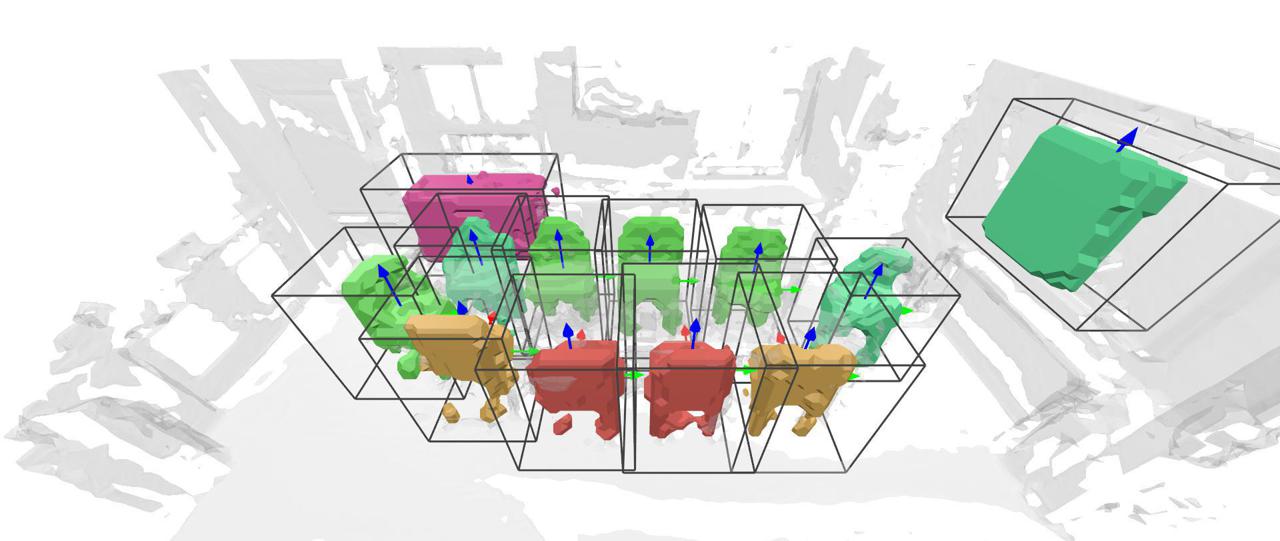}
		\includegraphics[width=\textwidth]
		{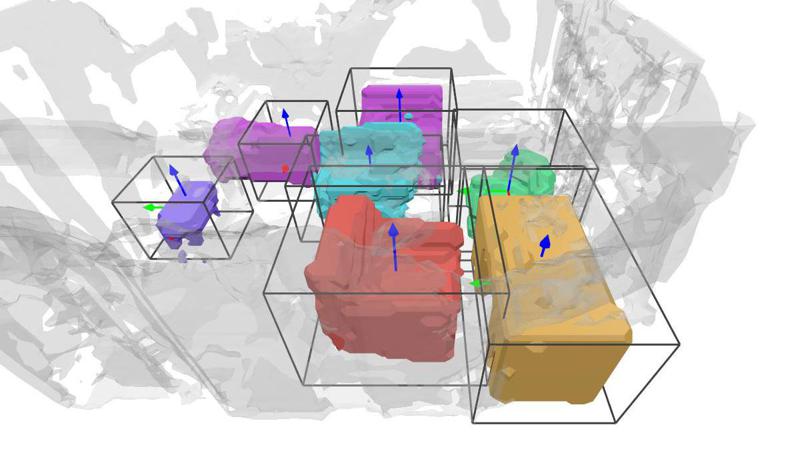}
		\includegraphics[width=\textwidth]
		{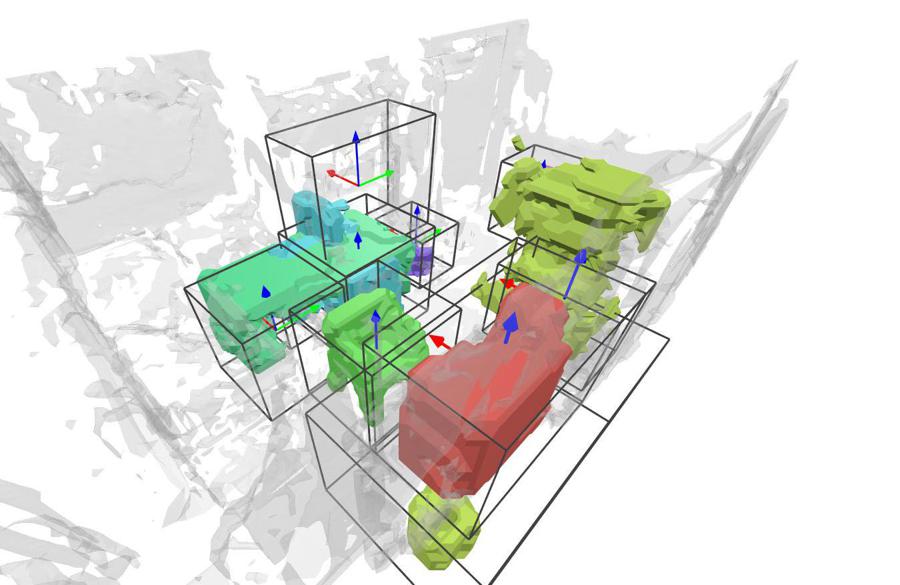}
		\includegraphics[width=\textwidth]
		{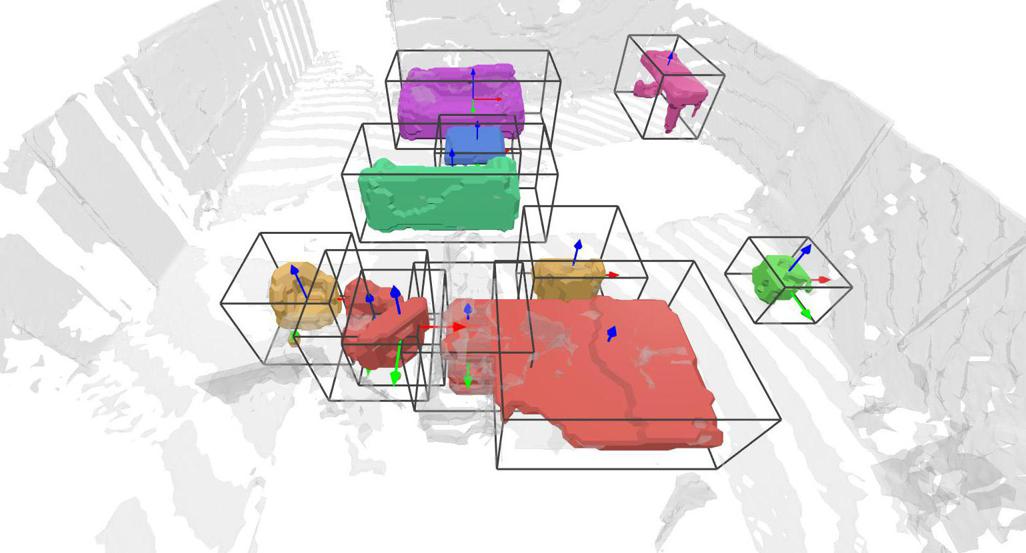}
		\includegraphics[width=\textwidth]
		{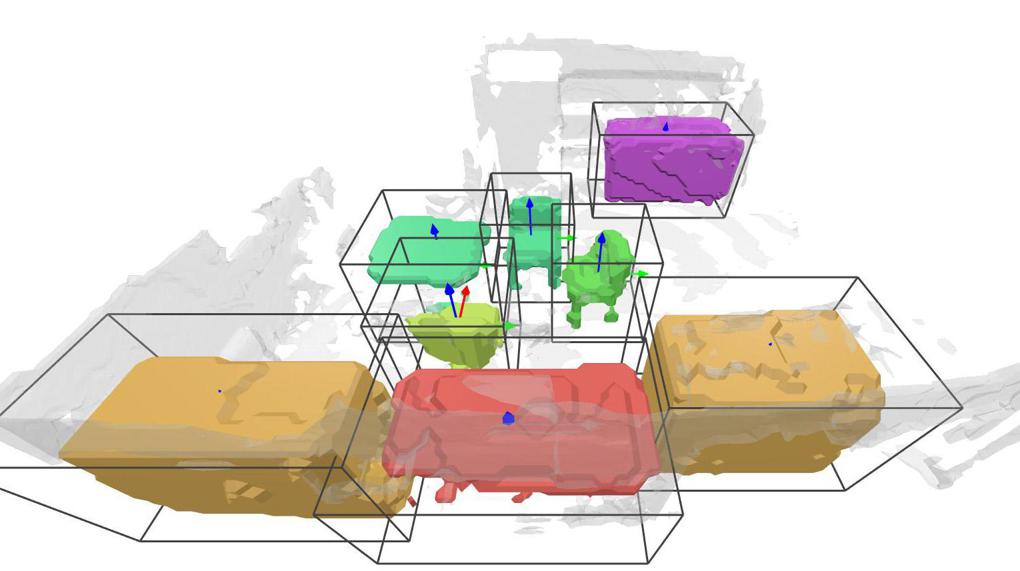}
		\includegraphics[width=\textwidth]
		{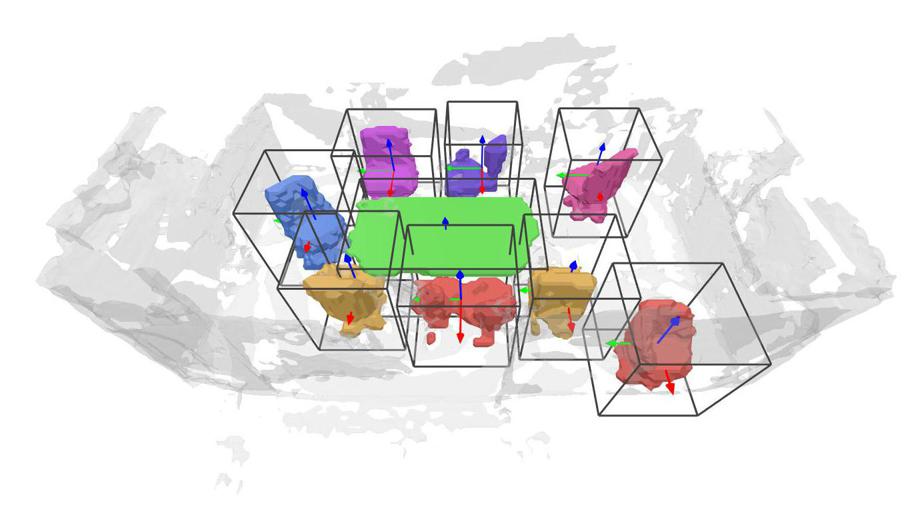}
		\includegraphics[width=\textwidth]
		{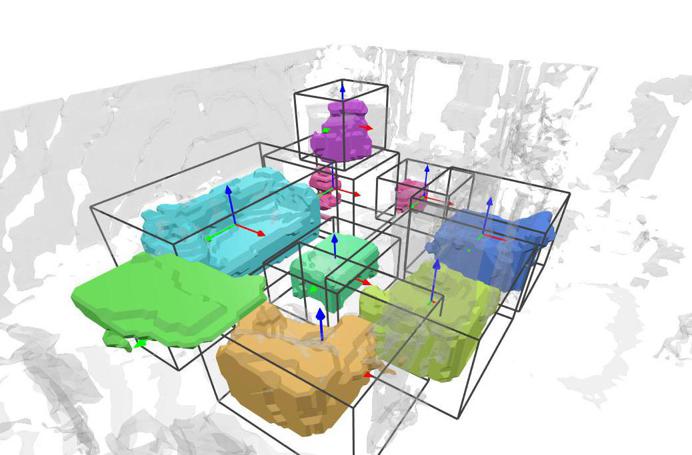}
		\includegraphics[width=\textwidth]
		{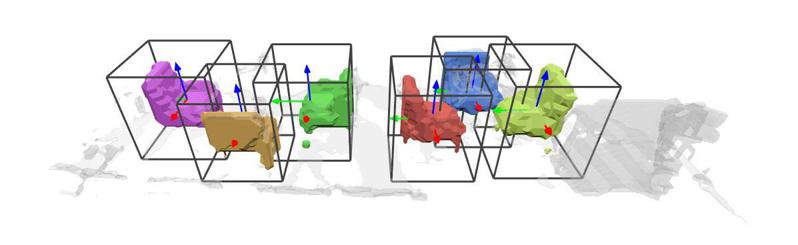}
		\includegraphics[width=\textwidth]
		{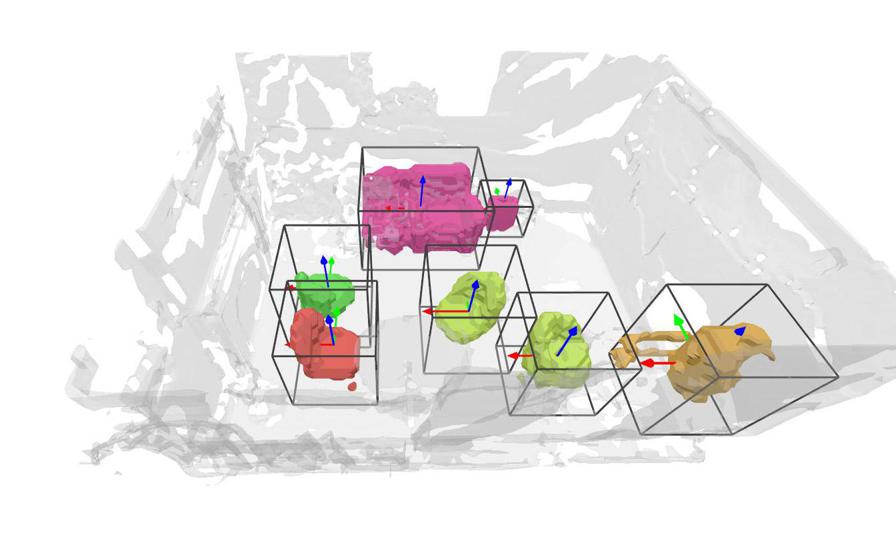}
		\includegraphics[width=\textwidth]
		{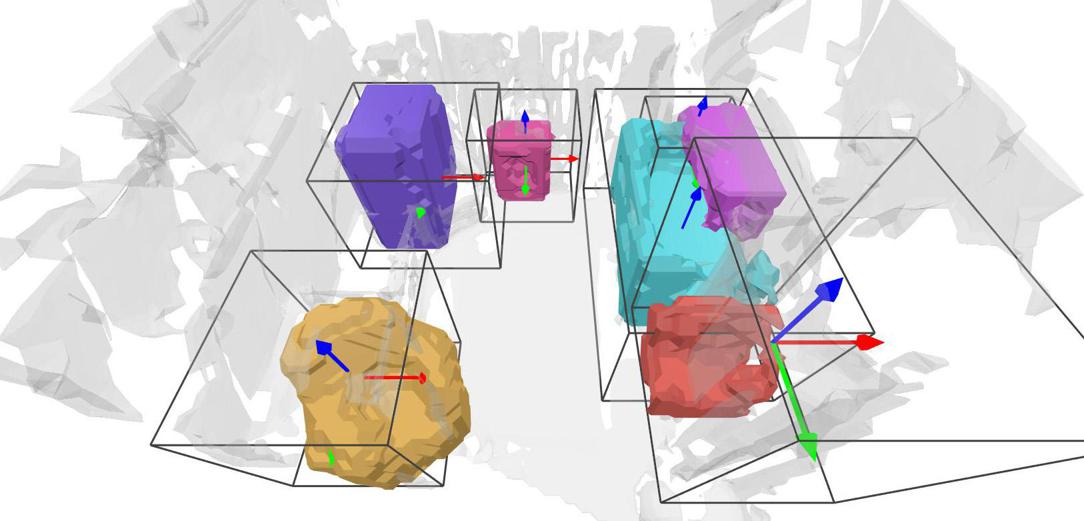}
		\caption{RevealNet \cite{hou2020revealnet} (Geo Only)}
	\end{subfigure}
	\begin{subfigure}[t]{0.24\textwidth}
		\includegraphics[width=\textwidth]  
		{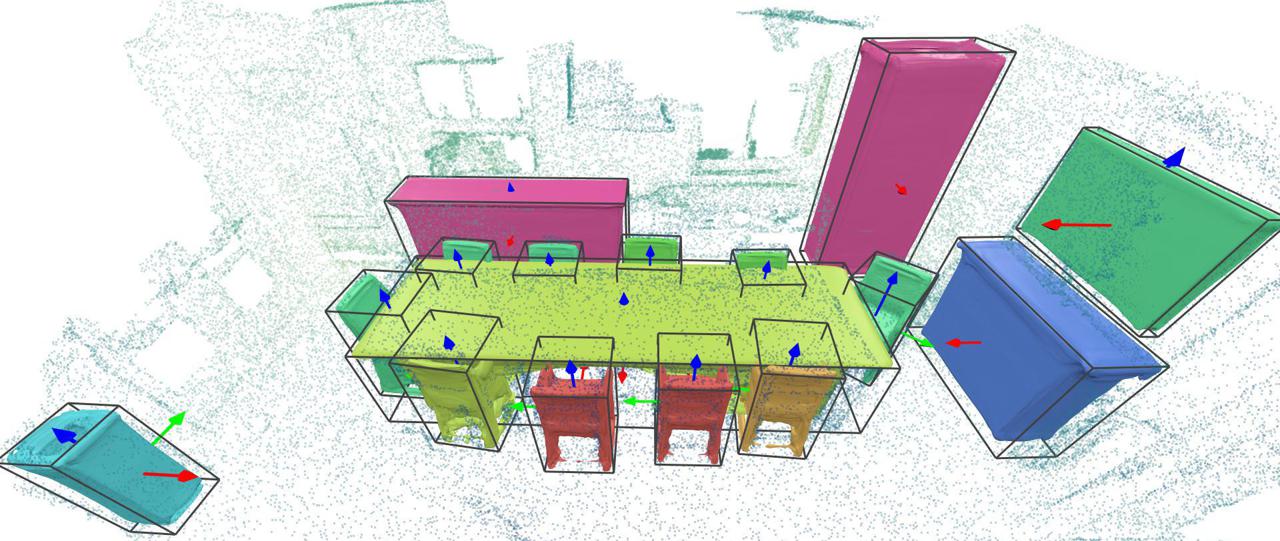}
		\includegraphics[width=\textwidth]  
		{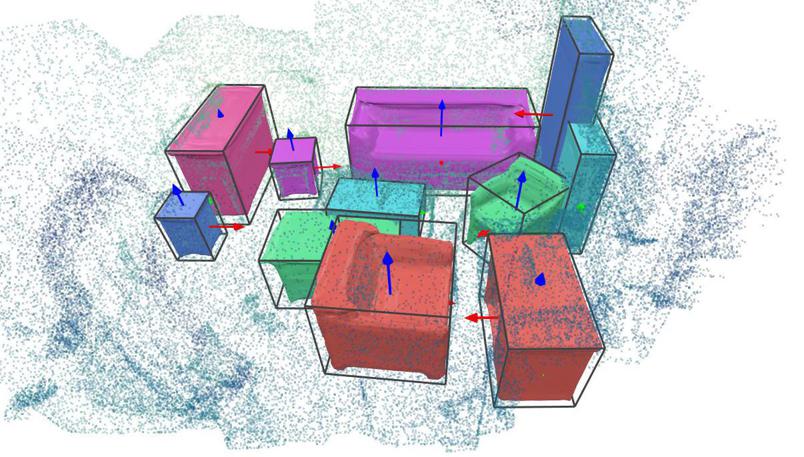}
		\includegraphics[width=\textwidth]  
		{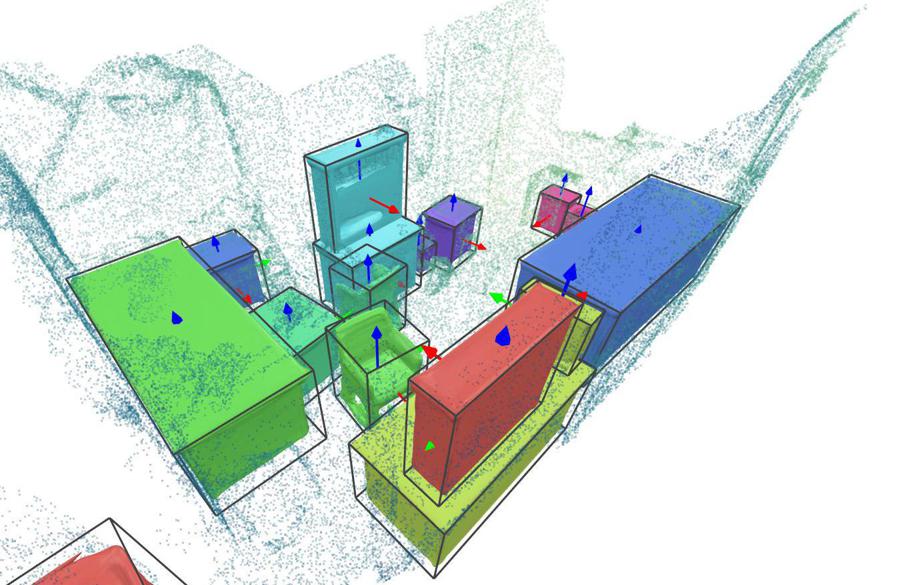}
		\includegraphics[width=\textwidth]  
		{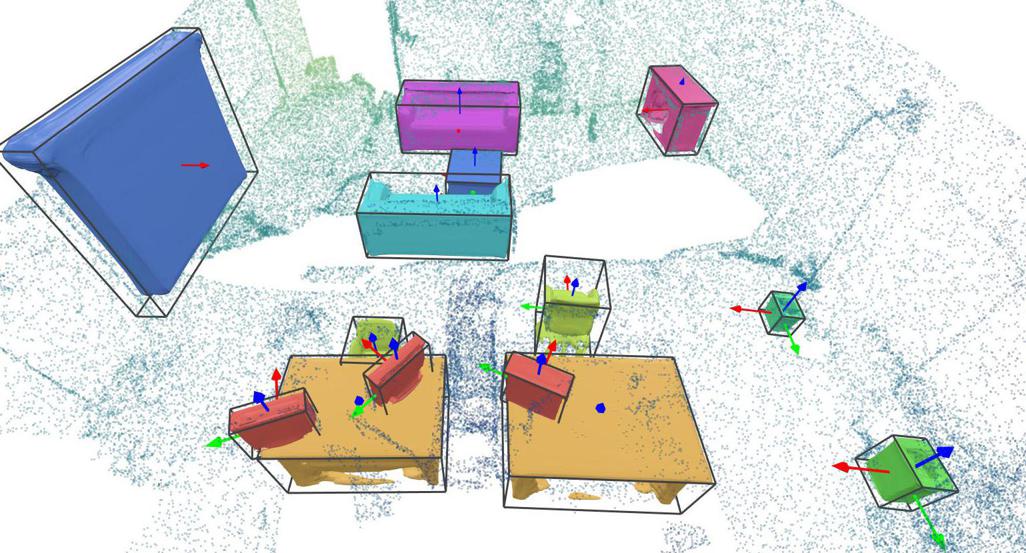}
		\includegraphics[width=\textwidth]  
		{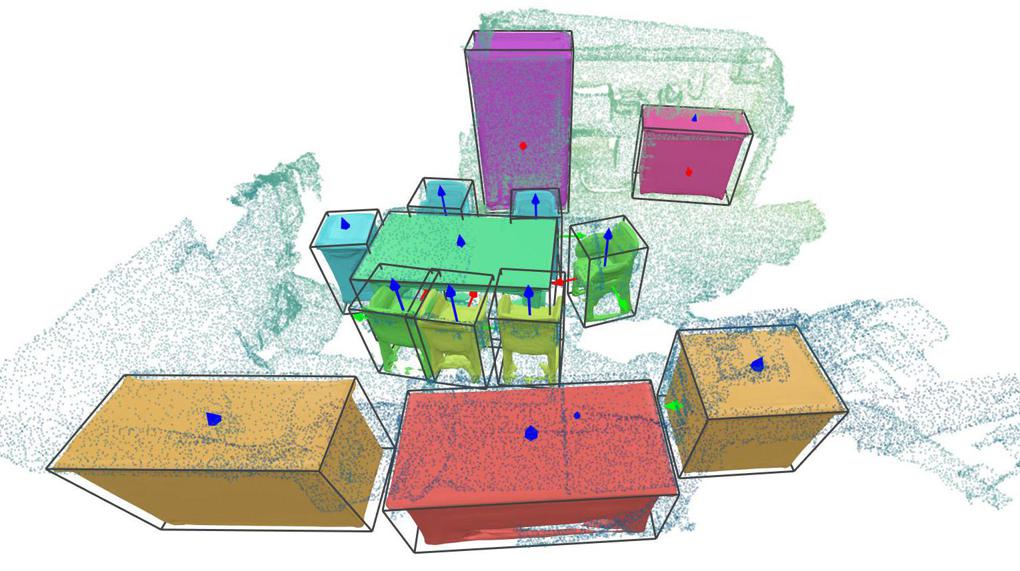}
		\includegraphics[width=\textwidth]  
		{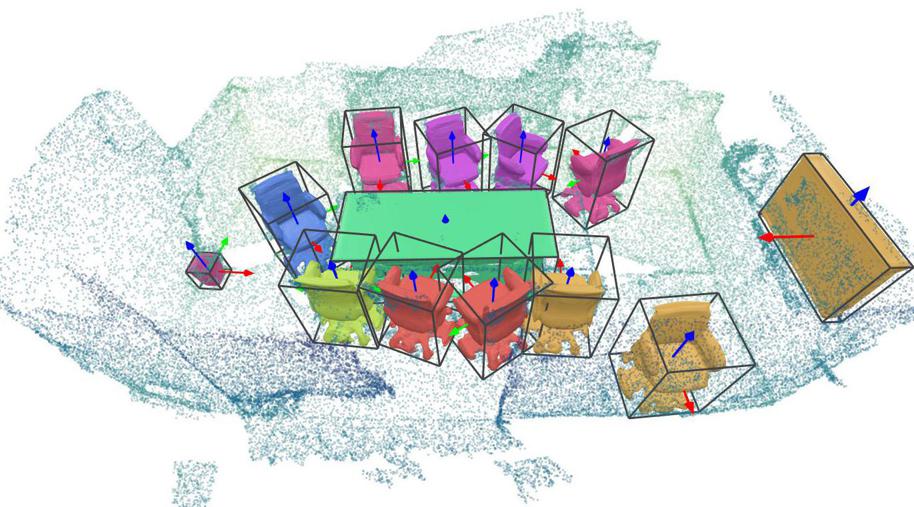}
		\includegraphics[width=\textwidth]  
		{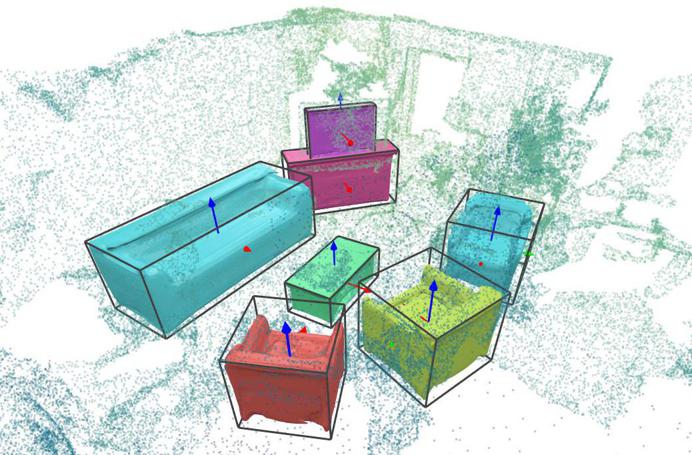}
		\includegraphics[width=\textwidth]  
		{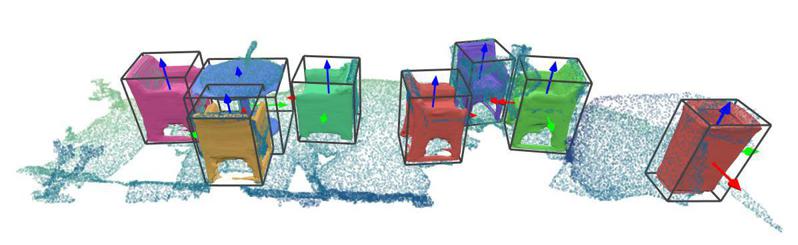}
		\includegraphics[width=\textwidth]  
		{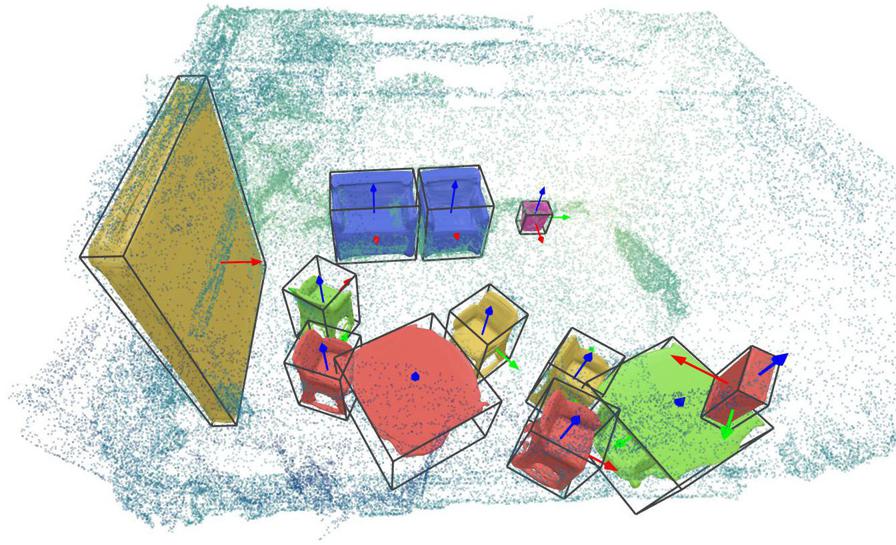}
		\includegraphics[width=\textwidth]  
		{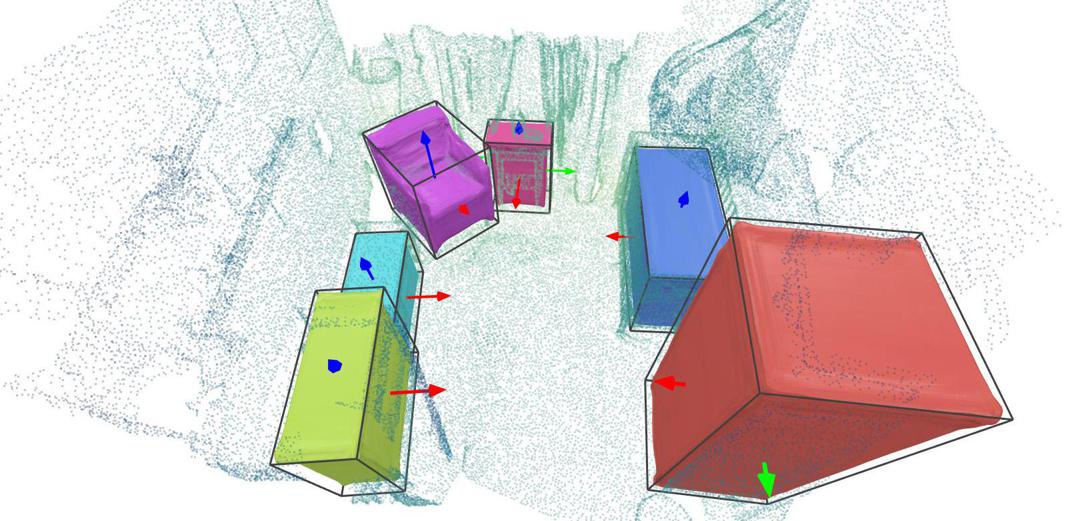}
		\caption{Ours (Geo Only)}
	\end{subfigure}
	\begin{subfigure}[t]{0.24\textwidth}
		\includegraphics[width=\textwidth]  
		{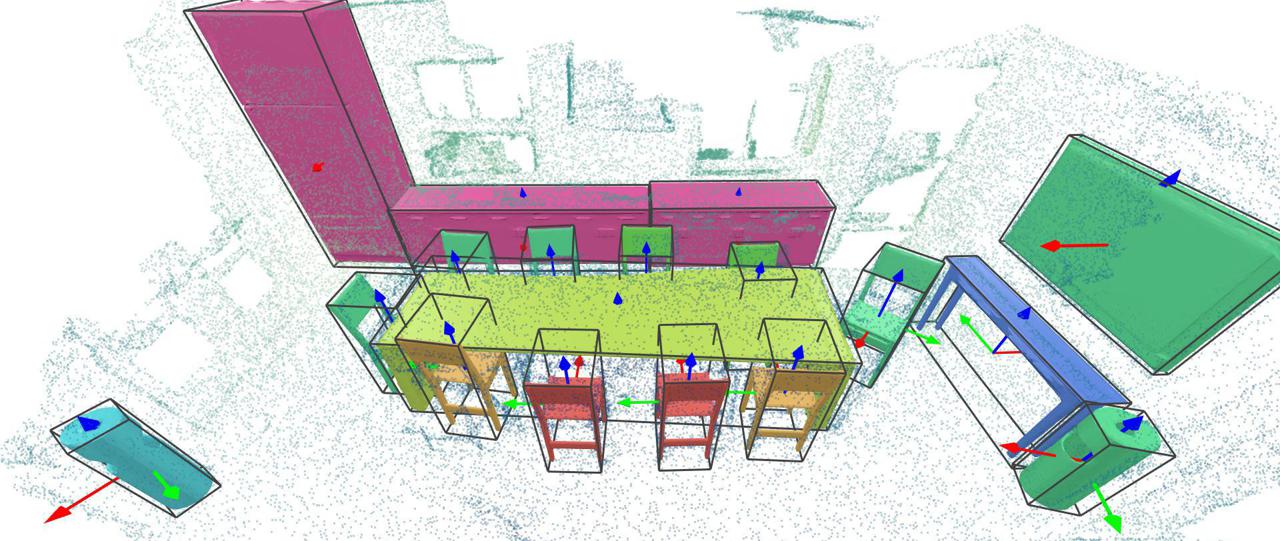}
		\includegraphics[width=\textwidth]  
		{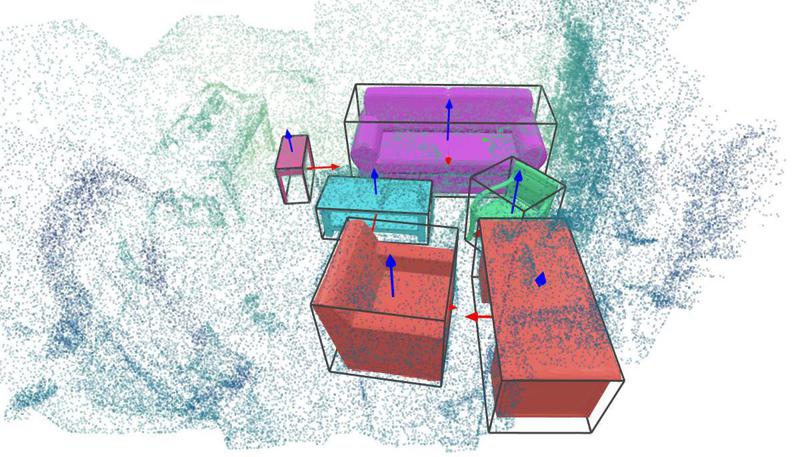}
		\includegraphics[width=\textwidth]  
		{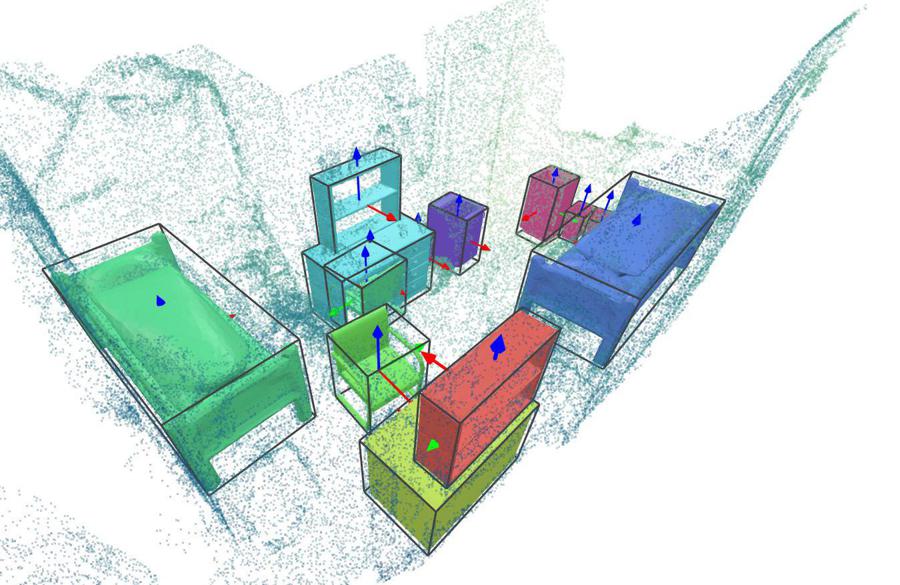}
		\includegraphics[width=\textwidth]  
		{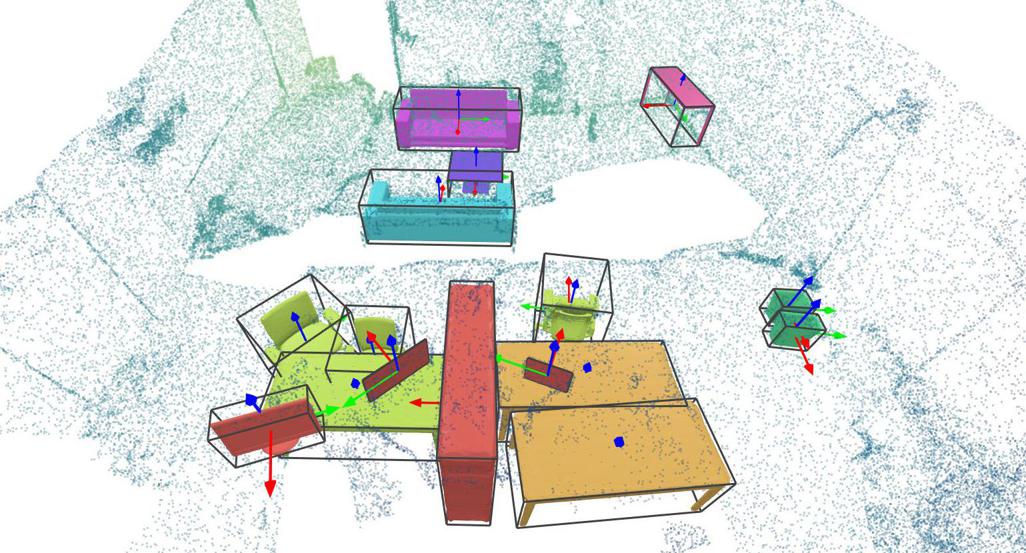}
		\includegraphics[width=\textwidth]  
		{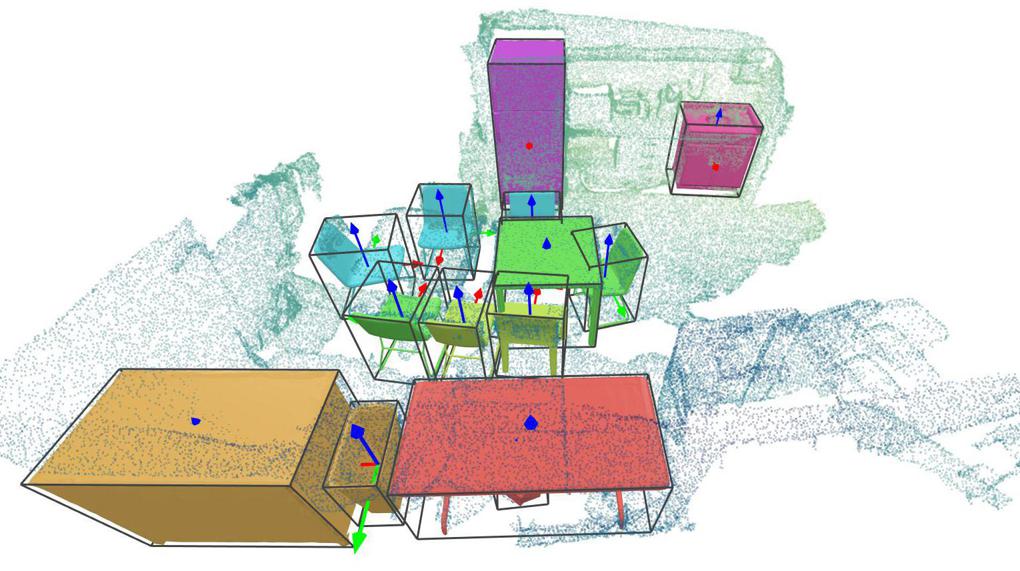}
		\includegraphics[width=\textwidth]  
		{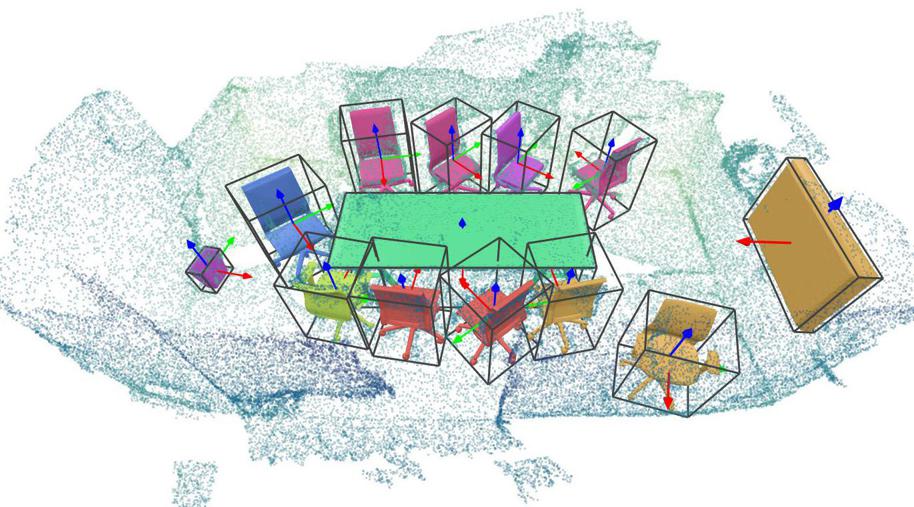}
		\includegraphics[width=\textwidth]  
		{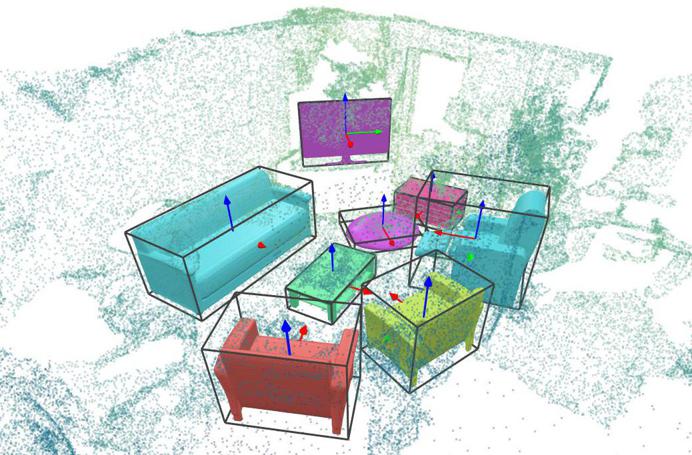}
		\includegraphics[width=\textwidth]  
		{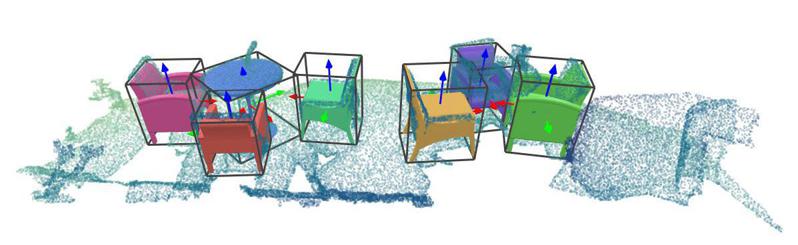}
		\includegraphics[width=\textwidth]  
		{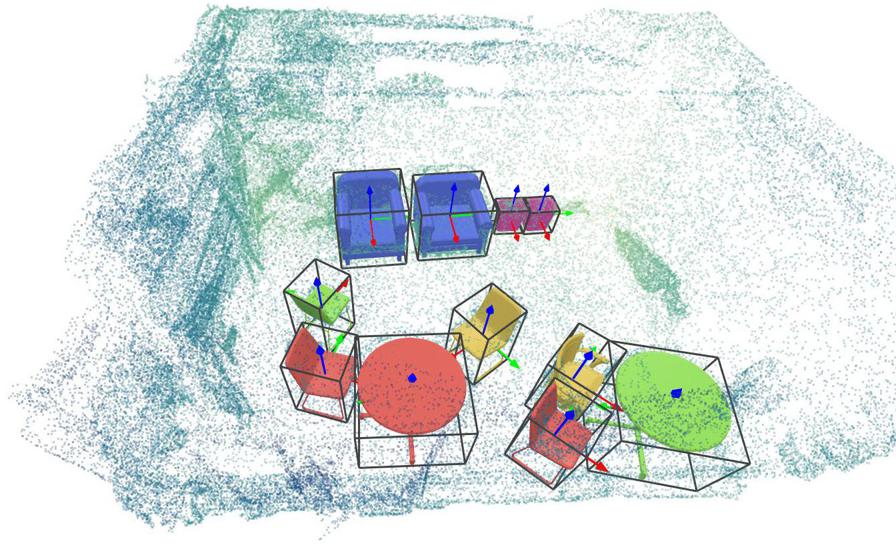}
		\includegraphics[width=\textwidth]  
		{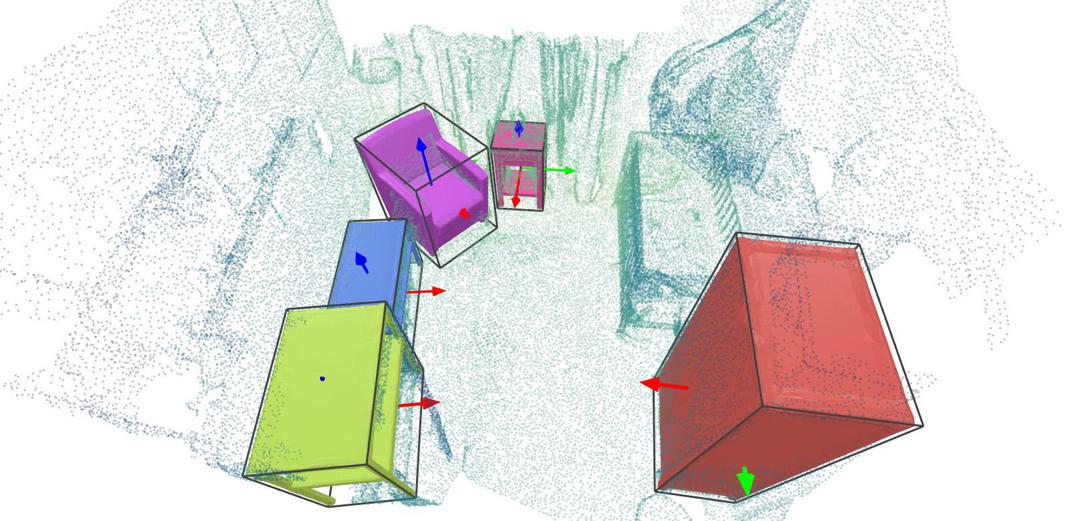}
		\caption{GT}
	\end{subfigure}
	\caption{Qualitative results of semantic instance reconstruction on ScanNet v2~\cite{dai2017scannet}. Note that RevealNet \cite{hou2020revealnet} preprocesses the scanned scenes into TSDF grids, while our method only uses the raw point clouds.}
	\label{fig:scene_recon_supp}
	\vspace{-10pt}
\end{figure*}

We provide more qualitative comparisons of semantic instance reconstruction with \cite{hou2020revealnet} in Figure~\ref{fig:scene_recon_supp}. As in RevealNet \cite{hou2020revealnet}, we use the groundtruth objects in Scan2CAD \cite{avetisyan2019scan2cad} for supervision. Note that the groundtruth CAD models in Scan2Cad are only partially labeled.

\end{appendix}

\end{document}